\documentclass[Afour,sageh,times,]{sagej}

\usepackage{moreverb,url}

\usepackage[colorlinks,bookmarksopen,bookmarksnumbered,citecolor=red,urlcolor=red]{hyperref}

\newcommand\BibTeX{{\rmfamily B\kern-.05em \textsc{i\kern-.025em b}\kern-.08em
T\kern-.1667em\lower.7ex\hbox{E}\kern-.125emX}}

\graphicspath{{./figs/}}
\usepackage{multicol}
\usepackage{amsmath} 
\usepackage{amssymb}  
\usepackage{float}
\usepackage{afterpage}
\usepackage{stfloats}
\usepackage{tabularx}
\usepackage{array}
\usepackage{makecell}
\usepackage{graphicx}
\usepackage{booktabs}
\usepackage{multirow}
\usepackage{dsfont}
\usepackage{romannum}
\usepackage{xcolor}
\usepackage{makecell}
\usepackage{enumitem}
\usepackage[format=plain, labelfont=it]{subcaption}
\usepackage{verbatimbox}

\newcommand*{\BeforeCaptionVSpace}{1ex}

\makeatletter
\newcommand{\labeltext}[3][]{%
    \@bsphack%
    \csname phantomsection\endcsname
    \def\tst{#1}%
    \def\labelmarkup{}
    \def\refmarkup{}%
    \ifx\tst\empty\def\@currentlabel{\refmarkup{#2}}{\label{#3}}%
    \else\def\@currentlabel{\refmarkup{#1}}{\label{#3}}\fi%
    \@esphack%
    \labelmarkup{#2}
}
\makeatother

\renewcommand{\eqref}[1]{(\ref{#1})}

\newcommand{\subfig}[1]{\textit{#1}}

\newcommand{\taskref}[1]{Task~\ref{#1}}

\newcommand{\ie}{\textrm{i.e.}}
\newcommand{\eg}{\textrm{e.g.}}

\def\intentionnet{IntentionNet}

\def\proposed{DECISION}

\def\human{Human Teleop}

\def\SR{SR}
\def\CR{Avg. Int.}

\def\intentforward{\texttt{go-forward}}

\def\inetlr{Kilo-IntentionNet}
\def\weblink{\urllink[pre = \bgroup\bf, post = \egroup]}
\newcommand{\cell}[1]{\#{#1}}

\newcounter{alphaenum}

\def\journalname{The International Journal of Robotics Research}
\def\volumeyear{xxxx}
\def\volumenumber{yy}
\def\issuenumber{zz}

\begin{document}
\renewcommand{\thepage}{\arabic{page}}

\runninghead{Wei Gao et al.}

\title{\intentionnet{}: Map-Lite Visual Navigation at the Kilometre Scale}

\author{Wei Gao\affilnum{1}, Bo Ai\affilnum{1}, Joel Loo\affilnum{1}, Vinay\affilnum{1} and David Hsu\affilnum{1}}
\affiliation{\affilnum{1}National University of Singapore}

\corrauth{Wei Gao, National University of Singapore.}
\email{a0134661@u.nus.edu}

\begin{abstract}

This work explores the challenges of creating a scalable and robust robot navigation system that can traverse both indoor and outdoor environments to reach distant goals. We propose a navigation system architecture called \intentionnet{} that employs a monolithic neural network as the low-level planner/controller, and uses a general interface that we call \textit{intentions} to steer the controller. The paper proposes two types of intentions, Local Path and Environment (LPE) and Discretised Local Move (DLM), and shows that DLM is robust to significant metric positioning and mapping errors. The paper also presents Kilo-\intentionnet{}, an- instance of the \intentionnet{} system using the DLM intention that is deployed on a Boston Dynamics Spot robot, and which successfully navigates through complex indoor and outdoor environments over distances of up to a kilometre with only noisy odometry.

\end{abstract}

\keywords{Autonomous navigation, Visual navigation, Deep learning}

\maketitle

\section{Introduction}
\label{sec:1}

\begin{figure*}[h]
    \centering
    \subfloat[]{
        \includegraphics[width=0.48\textwidth]{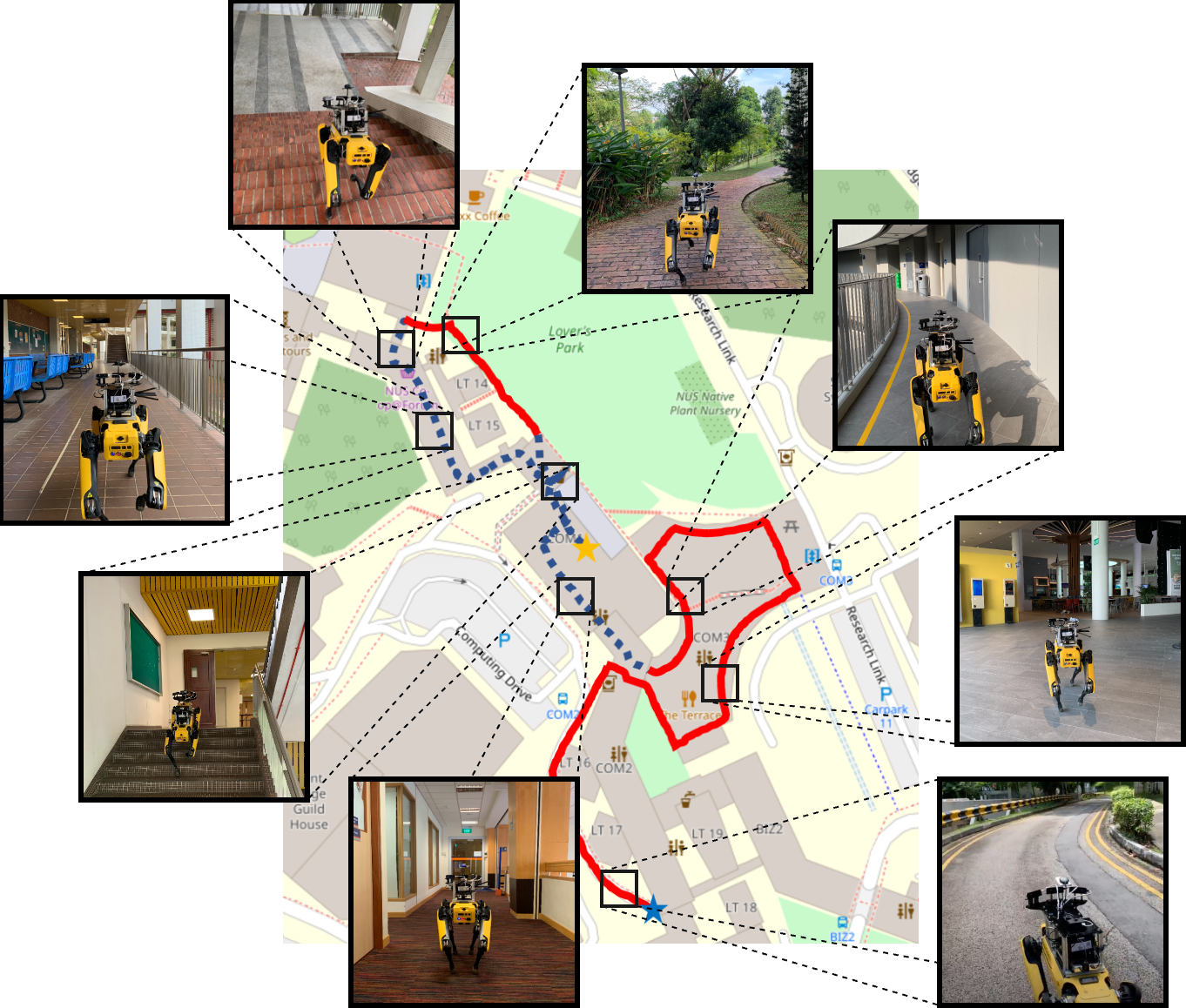}
        \label{subfig:long_range_route1}
    }
    \subfloat[]{
        \includegraphics[width=0.38\textwidth]{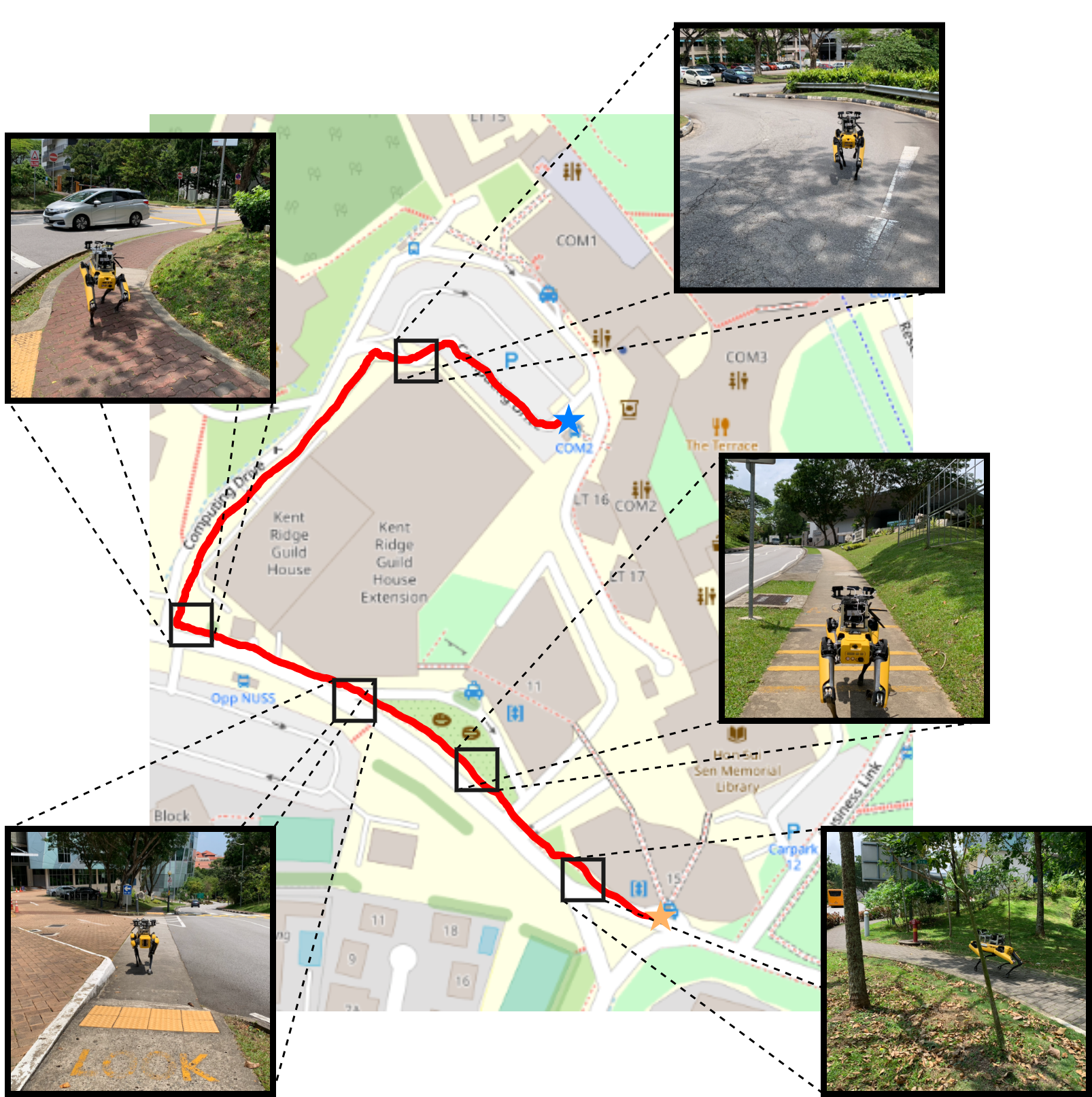}
        \label{subfig:long_range_route2}
    }
    \caption{We demonstrate \inetlr{}'s capability for long-range navigation on complex routes that mix diverse indoor and outdoor environments, and that cover distances of up to a kilometre. While \inetlr{} uses a learned controller, we show that it is capable of generalising to visually different environments not seen in its training data: the red segments of the path indicate novel environments which the controller was not trained on. The orange and blue stars mark the start and end of each route respectively.}
    \label{fig:long_range_routes}
\end{figure*}

How can we imbue robots with the ability to navigate through \textit{diverse environments} to \textit{distant goals}? This remains an open challenge due to the complexity and difficulty of designing a robot that can generalise over environments, tolerate significant mapping and positioning inaccuracies and recover from inevitable navigation errors. While many works tackle robot navigation, few systems capable of long-range, kilometre-scale navigation exist. Classical robot systems capable of long-range navigation like \cite{montemerlo_stanford_junior, kummerle2013obelix} use e xplicit maps and find paths over them using classical planning algorithms~\citep{siegwart_amr}, allowing them to reach arbitrarily distant goals in principle. However, they often require high mapping and positioning accuracy to work well, and being constructed from many handcrafted sub-components they can be brittle and difficult to tune in the real world. In contrast, many modern data-driven navigation approaches learn direct mappings from observations to actions~\citep{zhu_drl_nav, gupta_cmp} which enables robust performance in the real world. Such systems are typically not able to navigate long distances, as they have limited ability to generalise to environments far outside their datasets and sacrifice compositional generality for performance. 

In this work we combine elements from both the classical and modern data-driven approaches, with the goal of designing a robot navigation system which can be used to robustly traverse varied environments and scale up to long-range navigation tasks on the order of kilometres. Our proposed system design is a two-level navigation architecture in which the lower level handles control and local obstacle avoidance, and is guided along the global path to the goal planned by the upper level. Drawing from classical approaches, the upper level of our system design retains an explicit map and employs classical planning algorithms on the map for path-finding, enabling the system to plan long trajectories to reach distant goals. Inspired by modern data-driven approaches, the lower level of our system design is a neural network-based controller that maps observations directly to velocity commands, and which is learned end-to-end from real world experience. Neural networks have the flexibility to accept a wide variety of input types, and we find that design space for the signals used by the system's upper level to guide the lower level is large. We exploit this property to design several different types of guidance signals, which we call \textit{intentions}. We find that designing the appropriate intention imbues the navigation system with specific abilities, such as the ability to tolerate significant mapping and positioning inaccuracies.

Overall, we find that our system design principles bring \textit{scalability} and \textit{robustness} to robot navigation systems with 4 key advantages:

\begin{enumerate}[label=\textbf{A\arabic*},ref=\textbf{A\arabic*}]
    \item \label{a_composition} \textit{Scales up planning to distant goals} by using classical planning algorithms that exploit compositionality.
    
    \item \label{a_avoidance} \textit{Robust control and obstacle avoidance} by learning a low-level controller from real-world data.
    
    \item \label{a_novel} \textit{Robust generalisation to novel environments} enables navigation to \textit{scale across heterogeneous, varied areas}.
    
    \item \label{a_distant} \textit{Robustness to mapping and positioning inaccuracies} through careful intention design enables navigation to \textit{scale over long distances.}
\end{enumerate}

Marrying elements from classical and modern data-driven approaches lets our proposed system design overcome some of the most significant shortcomings of each approach. We validate the 4 advantages of our system design with extensive experiments. We further implement an instance of our proposed system design that we call \inetlr{}, and demonstrate its capability for scalable and robust kilometre-scale navigation in the real world, through diverse indoor and outdoor environments, in the presence of clutter, vegetation and dynamic objects like pedestrians.

\section{Related work}
\subsection{Classical navigation system architecture}
The design and implementation of robotic navigation systems is well-studied, with many proposed solutions. As \cite{xiao_navsurvey, siegwart_amr} discuss, decomposition and modularity is a defining feature of classical navigation system architectures. Many systems are hierarchically decomposed, most often into a two-level architecture comprising a \textit{global planner} subsystem that finds a coarse path from the current position to the goal, and a \textit{local planner/controller} subsystem that generates low-level motion commands to actuate the robot to a local subgoal. Each level or subsystem is often further decomposed into sequential modules, such as perception, control, reasoning or planning. In classical robot systems, these modules usually consist of handcrafted, model-based algorithms.

\cite{oleynikova_mav} provide a recent example of the classical approach to robot navigation. Their drone navigation system contains a global planner subsystem that plans a coarse path to the goal with A$^{*}$ search over a sparse, topological map of the environment. This coarse path is communicated as a series of waypoints to their local planner/controller subsystem, which builds a high-resolution representation of the local environment and uses it to find and track a kinodynamically feasible path to the next waypoint. Specifically, the local planner/controller uses the Voxblox signed distance field mapping~\citep{oleynikova2017voxblox} for environment representation, continuous-time trajectory optimisation~\citep{oleynikova2016loco} for local planning, and model-predictive control to issue the commands to actuate the robot. Both subsystems rely on ROVIO~\citep{bloesch2017rovio} for state estimation and positioning. As can be seen, each module in this system is a handcrafted and hand-tuned model-based algorithm. Many classical robot systems follow this general architecture, albeit with different handcrafted modules. Examples range from the navigation systems in self-driving vehicles like Stanford's Junior~\citep{montemerlo_stanford_junior} and MIT's Talos~\citep{leonard_mit_talos}, to those in agile drones~\citep{gao2019aggressiveflight}, to the standard ROS navigation pipeline~\citep{ros_move_base} which is now widely used across many robot systems.

As \cite{karkus_dan} note, the performance of such classical systems can be limited either due to imperfections in the way the system is decomposed, or due to the imperfect priors and models used in designing the algorithms in each module. They further show that end-to-end learning from real-world data can help to overcome these issues. In a similar manner, the \intentionnet{} architecture seeks to incorporate end-to-end learning in classical navigation systems to address these shortcomings.

\subsection{Learned navigation system architecture}
Many recent works have similarly proposed to incorporate learning into navigation systems. We draw on the taxonomy of \cite{xiao_navsurvey} to organise discussion of such works. We discuss works that either learn the entire navigation system or its subsystems, as they reflect an \intentionnet{} design principle that end-to-end learning can overcome imperfections from suboptimal decompositions. We refer the reader to the survey for in-depth discussion of other paradigms, such as learning individual modules in each subsystem.

An approach enabled by deep learning is to replace the entire robot navigation system with a monolithic neural network learned end-to-end. In such systems, the neural networks take in observations of the local environment along with a goal specification, and directly output control commands to actuate the robot toward the goal. Examples of this paradigm include \cite{pfeiffer_end2end_planning, zhu_drl_nav, gupta_cmp, sorokin_sidewalk_nav}. By learning from real-world data, such systems may perform more robustly in areas similar to their training data. However they lack compositional generalisation: they typically have limited ability to generalise to new environments or plan long trajectories that are not present in their training data distributions. \cite{pfeiffer_drl_il, pfeiffer_end2end_planning} show this limitation empirically and suggest that end-to-end methods should be primarily applied to navigating within a local area.

An alternative approach is to retain the hierarchical decomposition found in classical navigation systems. \cite{xiao_navsurvey} note that the vast majority of work in this area employ a handcrafted global planner with a learned local planner/controller. Such systems focus on learning only local skills such as obstacle avoidance, while sidestepping the need to learn long-range trajectories from data. They are able to improve the robustness of local planning and control through learning, while enabling compositional generalisation to long-range navigation by using classical planning algorithms. For example, PRM-RL~\citep{faust_prm_rl} learns a controller for short-range navigation to a specified waypoint with deep reinforcement learning, and couples this with a global planner that operates on a Probabilistic Roadmap~\citep{kavraki_prm}. \cite{francis_prm_autorl} further improve the robustness and performance of PRM-RL's controller with the AutoRL framework~\citep{chiang2019autorl}. To perform the complex task of robust, socially acceptable navigation, \cite{pokle_social_nav} learn a controller that takes in a variety of perception data ranging from LiDAR observations to nearby humans' trajectories. This learned controller is coupled with a Dijkstra-based global planner operating on a 2D grid map of the environment. Instead of specifying local goals as individual waypoints like PRM-RL, their controller is conditioned on the entire path from the global planner. ViNG and RECON~\citep{shah2021ving, shah_recon} couple a learned image goal-based controller with a Dijkstra-based global planner operating on a topological graph. In these systems, each node in the graph contains an image of that location, and the global planner specifies the next local goal to reach by conditioning the controller on the associated image of that node.

The \intentionnet{} architecture takes the latter approach and retains the hierarchical structure of classical systems. It employs a classical global planner to generalise to long-range navigation tasks. At the same time, it parameterizes the local planner/controller as a monolithic neural network that is learned end-to-end, to make it more robust and compensate for imperfect decompositions and inaccurate models. Our work further recognises that implementing local planners/controllers as neural networks enables them to accept goals specified in a wide variety of ways, and we explore this goal specification design space. We term the interface by which the global planner specifies goals to the local planner/controller as \textit{intentions}, and propose 2 different intention types: \textit{Local Path and Environment} (LPE) which is a rich goal representation that can enable fine-grained navigation in systems with high localisation and mapping accuracy, and \textit{Discretised Local Move} (DLM) which is the set of high-level driving directions \{\texttt{turn-left}, \texttt{go-forward}, \texttt{turn-right}\} that is suited for systems with poor localisation and mapping accuracy. We note similarities between our DLM intention and the work of \cite{cil}. While their work focuses only on learning a driving directions-based controller, we explore DLM's potential for enabling long-range navigation with inaccurate maps.

\subsection{Long-range navigation systems}
We consider the subset of navigation systems, both classical and learned, that are designed to be capable of long-range navigation. Stanford's Junior~\citep{montemerlo_stanford_junior}, MIT's Talos~\citep{leonard_mit_talos} and other self-driving vehicles entered into DARPA's Grand and Urban Challenges were designed to tackle on-road routes stretching over many kilometres. More recently, ViKiNG~\citep{shah2022viking} and ViNT~\citep{shah2023vint} combine topological graphs with visual navigation controllers to achieve kilometre-scale navigation over varied urban terrain with mobile robots. However, these systems largely assume the availability of accurate localisation (i.e. GPS) and accurate maps (e.g. road networks, geo-referenced satellite maps). These assumptions may not hold, especially in built-up urban areas where GPS signals are weak, or in dynamic, changing environments. While SLAM techniques can be used for mapping and localisation, ensuring good performance over long distances and diverse environments remains an open problem~\citep{cadena_slam} and recent works on scaling up navigation systems focus on overcoming this issue. Obelix~\citep{kummerle2013obelix} is a graph SLAM-based navigation system capable of travelling over 3km through a crowded city centre. However, it needs to have a highly accurate metric map for autonomous navigation and relies on GPS to obtain one. By learning its local planner/controller, PRM-RL~\citep{faust_prm_rl} achieves greater robustness to sensor and localisation noise that enables long-range indoor navigation. Similar to Obelix, it also requires accurate metric information to generate its PRM global map. RAVON~\citep{braun2009ravon} uses an abstract, coarse topological map for global planning and only relies on accurate metric information for local obstacle avoidance. They demonstrate navigation of up to 1km in unstructured outdoor terrain with their system. \cite{liu2021EfficientAR} propose a navigation system similar to our DLM-based \intentionnet{} design, that uses a learned controller to robustly navigate with the guidance of only noisy GPS signals and coarse-grained GPS maps. Their system is capable of on-road navigation over trajectories of more than 3km.

Our \inetlr{} system combines the \intentionnet{} architecture with DLM intentions to specifically compensate for inaccurate maps and localisation, and enable kilometre-scale navigation. In contrast with most of the above works, which focus on only either indoor or outdoor environments, our system is designed to navigate through \textit{both} indoor and outdoor environments with diverse visual appearances to reach a specified goal. It is able to achieve this by using both geo-referenced road networks and inaccurate metric maps like floor-plans to guide navigation.

\subsection{Learned controllers for visual navigation}
The \intentionnet{} architecture is premised on learning a visual navigation controller to replace the classical local planner/controller pipeline, which is a challenging task when using visually rich sensory inputs like RGB images. We note that in this work, we refer interchangeably to a learned local planner/controller subsystem as a low-level controller, to align with terminology in visual navigation literature.

A common way to train such a low-level controller is to use RL in simulation~\citep{faust_prm_rl, sorokin_sidewalk_nav, kaufmann_drone_racing}. However, these methods often face the sim-to-real issue, where the simulation-trained policy is unable to generalise to the real world~\citep{kendall_drive_rl, shai_multiagent_rl, sallab_drl}. While some works attempt to find ways to help simulation-trained policies generalise~\citep{loquercio_racing_sim2real}, others argue that direct experience of the real world is irreplaceable~\citep{levine_nav_exp}.

Methods that learn from real world experience have found notable success on real-world tasks. Imitation learning (IL) has been used for tasks including lane-following~\citep{pomerleau_alvinn, bojarski2016end}, collision avoidance~\citep{muller_obs_avoid} and high-speed drone flight in cluttered environments~\citep{loquercio_highspeed_flight}. Offline RL techniques that learn from static pre-collected datasets of robot trajectories~\citep{levine_offline_rl} have also been successfully applied to learn goal-reaching navigation policies that also satisfy auxiliary user-defined rewards~\citep{shah_revind}.

We posit that real-world experience is crucial for effectively learning a low-level controller that can both perform well in the real world and generalise to novel scenarios. As \cite{levine_nav_exp} note, it is challenging for human engineers to capture all variations of real-world scenarios, so handcrafted, heuristic controllers or controllers learnt from handcrafted simulators are likely to have limited or even unrealistic notions of traversability. Instead, they suggest that learning from real-world data yields a controller with richer and more accurate representations of traversability. In this work, we build on our prior work in \cite{ai_decision} and employ imitation learning on an extensive dataset of indoor and outdoor environments to train a robust low-level controller for visual navigation.

\section{System overview}
\label{sec:3}

In this section, we provide an overview of our proposed \intentionnet{} navigation system design. \intentionnet{} can be adapted to a range of applications from fine-grained, precise navigation to large-scale robust navigation. While we provide some discussion on the potential capabilities of our system design, we primarily focus on developing a concrete instantiation of \intentionnet{} targeted at solving the problem described in \autoref{sec:1}: that is, how to scale navigation to long distances and ensure robust performance even when traversing varied, complex indoor and outdoor environments.

\begin{figure}[h]
  \centering
  \includegraphics[width=.99\linewidth]{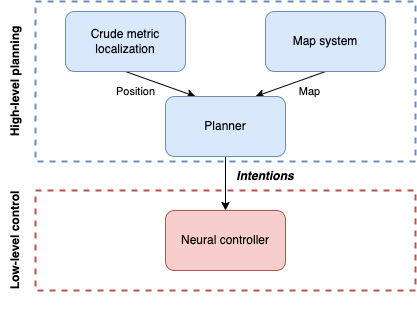}
  \vspace{\BeforeCaptionVSpace}
  \caption{Autonomous navigation system overview.}
  \label{fig:pipeline}
\end{figure}

\subsection{Architecture}

\intentionnet{} is a generic two-level navigation system design with a learned low-level controller and a general interface that we call \textit{intentions}, that is designed to connect a classical high-level planner with a learned low-level controller. \autoref{fig:pipeline} provides an overview of the system architecture.

To handle visual complexity, perceptual uncertainty and navigate robustly in a wide range of environments, \intentionnet{} learns a low-level controller that directly maps visual observations to control outputs. Trained with imitation learning (IL), this low-level controller implicitly learns collision avoidance and path following skills and generalises to a wide range of environments. Since our system makes only weak assumptions on the accuracy of the map provided to the high-level planner, it is essential to have a low-level controller that has robustly learned collision avoidance and does not assume the high-level planner will give specific guidance on how to avoid obstacles locally. Further details on the design principles and implementation of our low-level controller can be found in \autoref{sec:4}.

\intentionnet{} adds a hand-engineered high-level planner and map system on top of the low-level controller to reach faraway goals. While the controller can only navigate in local regions, the high-level planner enables compositional generalisation by finding a coarse global path, then computing and issuing the necessary `steering' signals to guide the low-level controller along a local segment of this path. The design of our high-level planner and map system is detailed in \autoref{sec:5}.

We call the `steering' signals that the high-level planner communicates to the low-level controller \textit{intentions}. Exploiting neural networks' input flexibility, we design two kinds of \textit{intentions} and discuss the capabilities they enable in \autoref{sec:intentions}.

We specifically develop a concrete instantiation of the \intentionnet{} system for long-range robust navigation at the kilometre scale, that we call \inetlr{}. The key observation behind \inetlr{} is that while environments can change, or while our perception of the environment can be affected by metric positioning and mapping error, the underlying structure or \textit{topology} of the environment usually remains constant. \inetlr{} is designed to recognise an environment's underlying \textit{topology} and navigate robustly even with noisy positioning or in dynamic environments. It does so using an intention interface that is a coarse discretisation of direction, and learns a low-level controller that can follow these intentions.

\subsection{Intentions}
\label{sec:intentions}
Intentions are `steering' signals by which the high-level planner guides the learned low-level controller along a global path. The low-level controller is conditioned on the intention from the high-level planner at each timestep, and the specific intention issued influences the control outputs. Intentions capture information about \textit{where} the robot should go in a local region, and \textit{how} it should travel there.

Since our low-level controller is parameterized by a neural network, the representation of the controller's intention input can take a wide variety of forms. We discuss two kinds of intention representations:

\begin{figure}
  \centering
  \includegraphics[width=0.5\columnwidth]{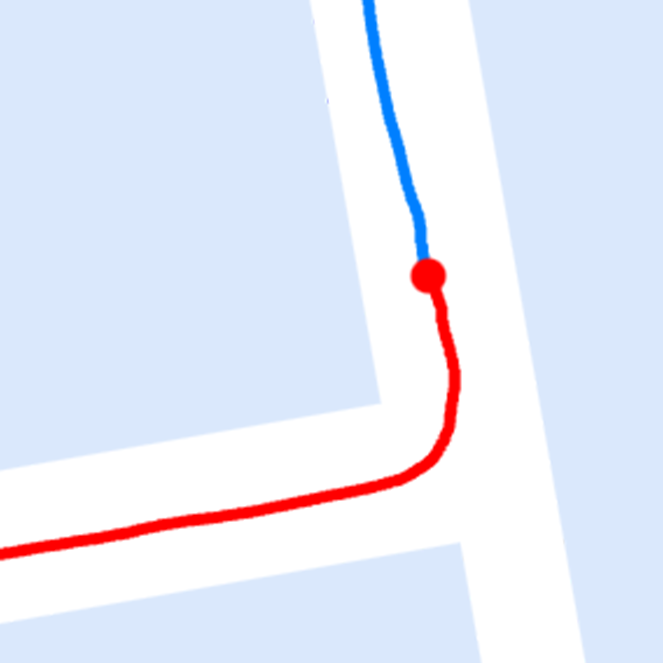}
      \caption{An illustrative representation of the Local Path and Environment (LPE) intention. It is a cropped section of the map, where the robot's historical path is visualized as a continuous red curve, while the planned future trajectory is delineated by a distinct blue curve. }
      \label{fig:LPE}
\end{figure}

\textbf{Local path and environment (LPE). }
LPE is a crop of a 2D top-down map that is centred on the current robot's position, and which is overlaid with the robot's motion history as well as the desired future trajectory the robot should take. \autoref{fig:LPE} provides an example of an LPE intention. LPE is a rich representation that includes some environmental information together with fine-grained information about the path the robot should take. A downside of this level of detail is that it requires high precision and accuracy in the robot's positioning and localisation to correctly overlay the path on the map.

\textbf{Discretised local move (DLM). }
DLM consists of a set of four discrete high-level commands that can be issued to the controller: \{\texttt{turn-left}, \texttt{go-forward}, \texttt{turn-right}, \texttt{stop}\}. DLM takes inspiration from driving/walking directions, which are largely capable of providing clear guidance for navigating in human-built environments, despite consisting primarily of these four types of commands. Since DLM commands represent a very coarse set of manoeuvres, precise and exact motion is not possible with them. DLM is also inadequate in environments where these commands become ambiguous (e.g. in a large open space like an amphitheatre), or in complex areas that require commands finer-grained than DLM's to traverse (e.g. at a junction with $\geq 5$ branches). On the flip side, the coarseness of DLM means that reliance on accurate positioning and localisation is somewhat reduced in systems that use this type of intentions. In addition, DLM can enable better shared autonomy, in the scenario that the high-level planner is replaced by a human user. Humans are more likely to find a semantically meaningful and coarse set of commands to be easier to use and more natural as compared to having to teleoperate the robot by issuing continuous velocity commands.

Intentions can vary widely in terms of the amount of information they contain. The LPE representation contains highly detailed and metrically accurate information, enabling a robot to navigate with high precision. While the simpler DLM intention contains no environment information and discards detailed path information, this simplified representation has less dependence on metric accuracy and can enable more robust navigation in the presence of noisy localization. The widely used `steering' input of a path specified by metric waypoints falls between LPE and DLM in terms of information content. Like LPE, such a path requires relatively high accuracy in mapping and positioning so as to provide a detailed and precise outline of the route to be traversed. Unlike LPE however, the path does not contain any information about the structure of the environment.

When navigating over long distances through complex environments, most robot systems will likely experience large amounts of localisation drift leading to metric inaccuracies in their positioning. Since LPE has a greater dependence on positioning accuracy than DLM, it is a less suitable intention representation for long-range navigation. While DLM sacrifices the ability to execute exact movements, it can work without absolute positioning and tolerate large amounts of localisation drift. We thus focus on incorporating DLM into \inetlr{} for the purpose of long-range navigation.

\section{Low-level controller}
\label{sec:4}
A key goal for the \intentionnet{} low-level controller is that it should enable \textit{robust control and obstacle avoidance} capabilities. To achieve this, we build the low-level controller based on two principles: 1) the controller should be a learnable, monolithic mapping from observations to controls, 2) the controller should be learned from real-world data. (1) enables the controller to learn to directly extract and use salient information for control from complex and rich observations like RGB input, reducing the brittleness of modular designs. Next, instead of depending on handcrafted and imperfect models of the world and of the robot's dynamics, (2) ensures that the models implicit in the controller are more realistic and accurate.

In this section, we describe the general architecture of an intention-steered controller, then discuss modifications that can be made to it for more robust performance in the presence of partial observability. Finally we describe how this architecture and design features are incorporated into the DECISION DLM-steered controller, which is our implementation of the low-level controller for \inetlr{}.

\subsection{Steering controllers with intentions}
\label{sec:controller_steering}
\begin{figure}
    \centering
    \subfloat[LPE]{
        \includegraphics[width=0.8\linewidth]{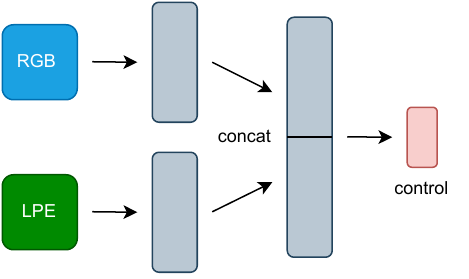}
        \label{subfig:lpe_arch}
    } \\ \vspace{10pt}
    \subfloat[DLM]{
        \includegraphics[width=0.8\linewidth]{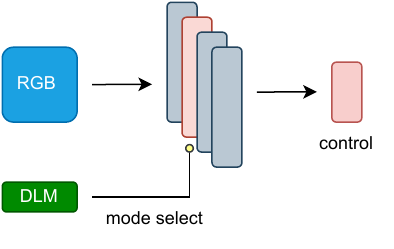}
        \label{subfig:dlm_arch}
    }
    \caption{Neural network controller architectures for different intention types. \subfig{(a)} LPE contains rich semantic information that can be extracted with CNNs. We directly concatenate LPE features and RGB features, since their information content is comparable. \subfig{(b)} DLM is a piece of symbolic information. Instead of concatenating it with the RGB features, we incorporate it into a switch module that conditionally selects the corresponding modes in the control predictions. }
    \label{fig:intention_archs}
\end{figure}

We describe the general setup and structure of an intention-steered controller, and give examples on how the controller can be specialised to use either DLM or LPE intentions.

\subsubsection{Controller backbone} Both our DLM- and LPE-steered controllers use a similar backbone to map input visual observations into output control commands. The first stage of the backbone is a CNN that takes RGB images as input and encodes them into latent image features. The latent features are passed through an MLP which outputs normalised linear and angular velocity values. These normalised control outputs are later scaled by a pre-selected maximum velocity, and the scaled velocity is tracked by the robot. We choose to use RGB inputs because they can capture textural cues that can highlight important traversability information (e.g. grassy terrain vs dirt path) which other modalities like depth may be unable to. This allows our controller to potentially capture a richer representation of traversability.

To capture real-world visual richness and complexity, we train the controllers to imitate expert trajectories collected in a diverse range of indoor and outdoor real-world environments with clutter and moving objects. 

\subsubsection{LPE-steered controller} Since the LPE intention also takes the form of an RGB image input, we encode it into a latent LPE image feature with a CNN backbone. The intention is injected into the controller backbone simply by concatenating the latent LPE image features with the latent image features from the controller's backbone, as shown in \autoref{subfig:lpe_arch}.

\subsubsection{DLM-steered controller} One possible way to condition the controller backbone on a categorical variable like the DLM intention is to represent it as a one-hot encoding, then map this into a latent feature that can be concatenated with the latent image features in the backbone. However we find empirically that this simple approach does not yield good performance, as each of the high-level behaviours specified by DLM produce substantially different actions and the single MLP in the backbone has difficulty learning the different behaviour modes simultaneously. In particular, learning a conditional multimodal policy in this manner can lead to mode collapse, yielding a unimodal policy whose action outputs are averaged across the different behaviour modes.

We overcome this by splitting the network into different parts, each specialising in a separate high-level behaviour mode. \autoref{subfig:dlm_arch} highlights this design principle: we learn separate MLPs for each behaviour, and use the DLM intention as a switch that selects the appropriate MLP for backpropagation/inference at train/run time.

\subsection{Improving robustness to partial observability}

While we demonstrated the effectiveness of the abovementioned LPE- and DLM-steered controllers in \cite{inet}, these controllers still have shortcomings, in particular a limited ability to handle partial observability.

Partial observability is a major issue in real-world navigation, often arising from the limited field of view of a robot's sensors, occlusions, dynamic obstacles, etc. Retaining temporal information can help us to overcome the partially observable nature of an instantaneous observation. We incorporate \textit{memory} into the controller architecture to maintain a more complete picture of the robot's surroundings when navigating. We modify the basic feed-forward controller backbone described in \autoref{sec:controller_steering} to include \textit{memory modules} and to make use of \textit{multiscale temporal modelling}.

\subsubsection{Memory module}
We use a modified Convolutional LSTM (ConvLSTM) \citep{convlstm} to aggregate spatio-temporal information. ConvLSTM is a variant of peephole LSTM  \citep{lstm_peep} where the input and output are 2D feature maps instead of 1D feature vectors. We additionally incorporate group normalization layers \cite{group_norm} and dropout to ease optimization and improve generalization. Details on the memory module structure can be found in \autoref{decision:architecture}.

\subsubsection{Multiscale temporal modelling}
We propose to apply this memory module to aggregate history information at multiple abstraction levels in the controller, to more effectively account for partial observability.

It has been shown that the earlier layers of a CNN tend to contain smaller-scale, low-level geometric features like edges and textures~\citep{convlstm, fcn, unet}, while later layers learn more abstract, larger-scale features useful for higher-level tasks like object detection~\citep{fpn, s3d}. 

We hypothesise that it is essential to use both low-level and high-level features for effective navigation. Detecting smaller-scale, low-level features allows us to accurately capture fine details like small movements or texture changes in the distance. Since higher-level features tend to be related to higher-level concepts of objects or shape, such information can be critical for a macro-level understanding of the environment and the interactions within it.

In practice we can implement this by using multiple memory modules in the network, where each module aggregates history information for a particular layer in the network, and fuses the history with that layer's features at the current timestep to enrich the representation.

\subsection{DECISION controller}
\begin{figure*}[t]
  \centering
  \includegraphics[width=\textwidth]{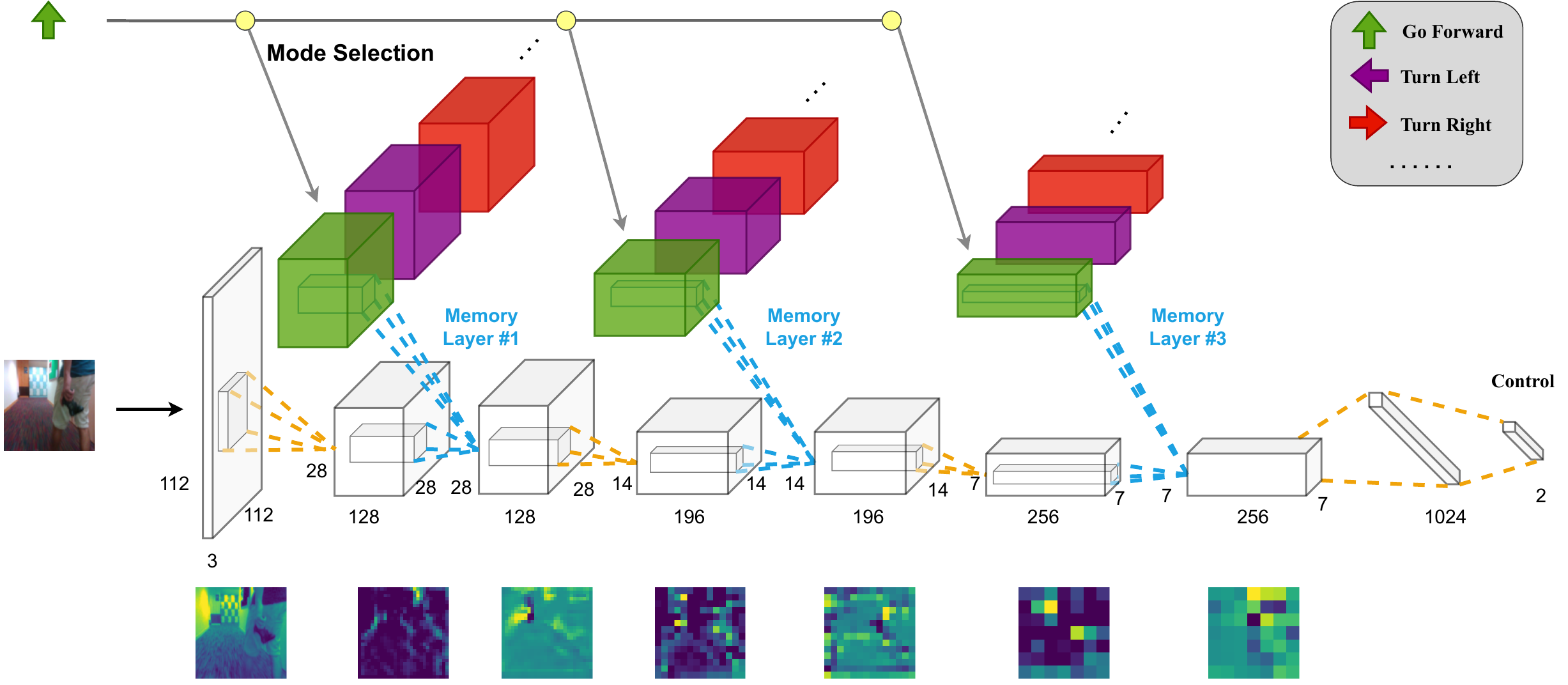} \\
  \caption{The neural network architecture of our \proposed{} controller. Every 3D volume in the figure denotes a feature map of shape (channel, width, height). The colored volumes denote the latent representation of the history, and blank volumes denote the spatial features at the current time step. Dashed lines between volumes denote operations such as convolutions. Key ideas: As features propagate through the convolutional layers, the representation becomes more abstract (visualized in bottom row), and the memory layers (\cell{1}-\cell{3}) integrate history information at multiple abstraction levels to enrich the representation. To learn multimodal behaviors, the memory modules for different modes in each memory layer are disentangled (volumes in different colors) and a symbolic signal is used to select the corresponding memory for feature propagation. }
  \label{archi}
\end{figure*}

We give an overview of the DECISION controller, which incorporates all the design features discussed above to enable robust control and obstacle avoidance. We design DECISION to be employed in \inetlr{} for scalable long-range navigation, and implement DECISION as a DLM-steered controller since DLM is a more suitable intention for such applications (\autoref{sec:intentions}). More details on the DECISION controller, as well as experiments validating its design features are provided in ~\cite{ai_decision}.

\subsubsection{Architecture of DECISION}
\autoref{archi} shows the architecture of the DECISION controller. To be robust to partial observability and aggregate spatiotemporal information, we employ our proposed memory module in DECISION. To capture temporal information at each feature map `scale', we design a network block that takes a feature map as input, fuses history information into it with our memory module, then transforms the fused output into a new feature map using a convolutional layer. Drawing on our earlier observations that different DLM behaviour modes are best learned using independent networks, we maintain a separate memory module for each behaviour. During both training and testing the DLM intention is provided to each network block, which uses it to select the corresponding memory module to use. Overall, each block transforms the input feature maps into an output set of smaller, higher-level feature maps that also incorporate temporal information from the currently executing intention.

The DECISION controller comprises 3 such network blocks in sequence, followed by an MLP that maps the flattened feature map into the normalised control command.

\subsubsection{Training DECISION}
\label{sec:learn}
\begin{figure*}[ht]
  \centering
  \setlength{\tabcolsep}{1pt} 
    \renewcommand{\arraystretch}{1} 
  \begin{tabular}{ccccc}
   \includegraphics[width=0.19\linewidth]{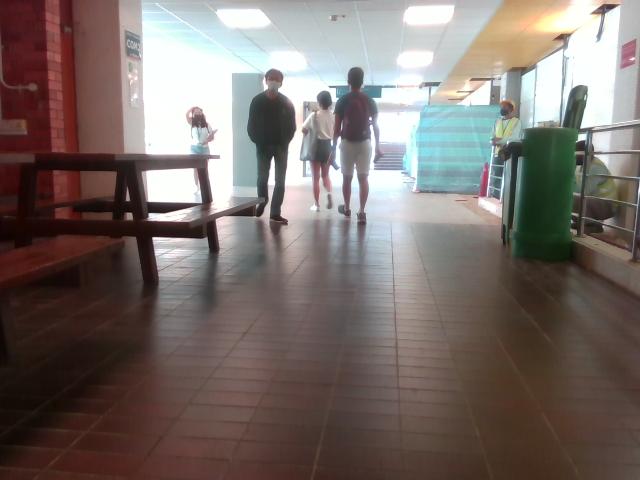} &
    \includegraphics[width=0.19\linewidth]{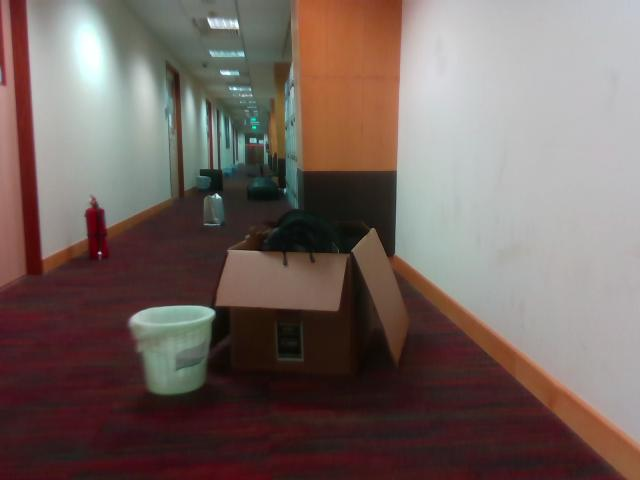} &
    \includegraphics[width=0.19\linewidth]{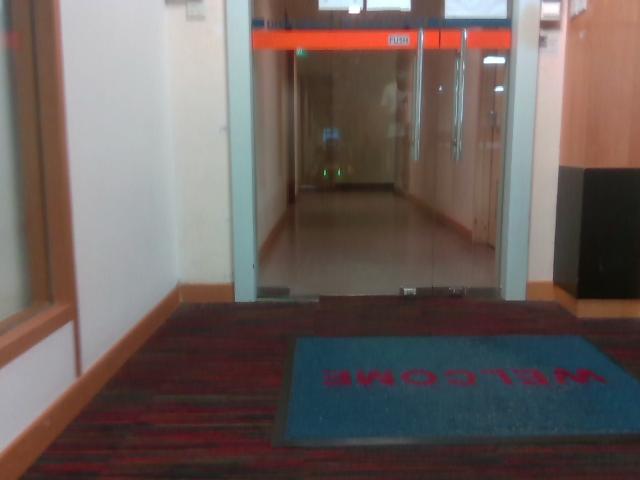} &
    \includegraphics[width=0.19\linewidth]{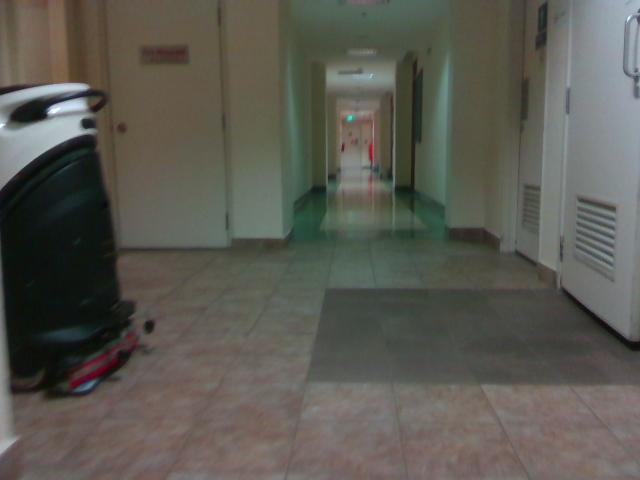} 
    \includegraphics[width=0.19\linewidth]{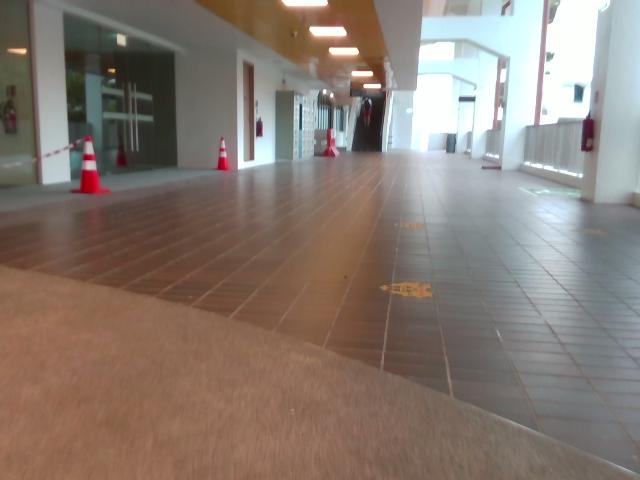} \\
    \includegraphics[width=0.19\linewidth]{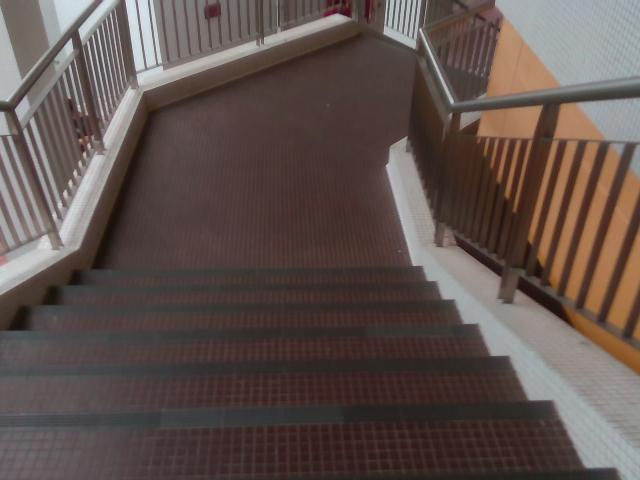} &
    \includegraphics[width=0.19\linewidth]{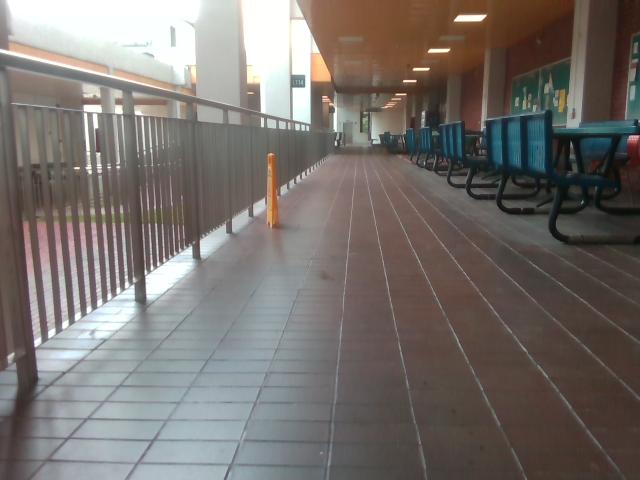} &
    \includegraphics[width=0.19\linewidth]{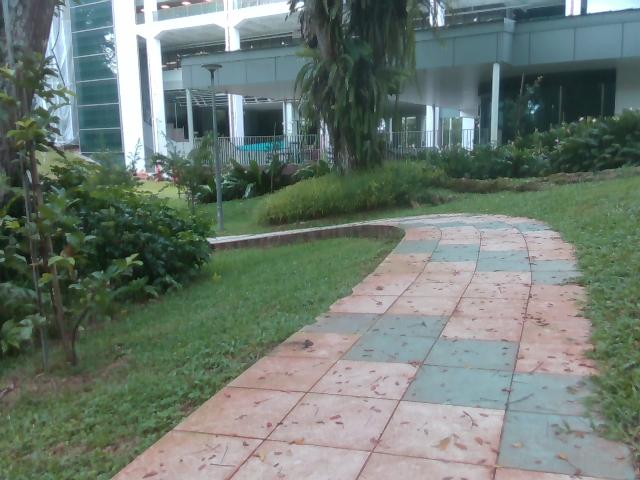} &
    \includegraphics[width=0.19\linewidth]{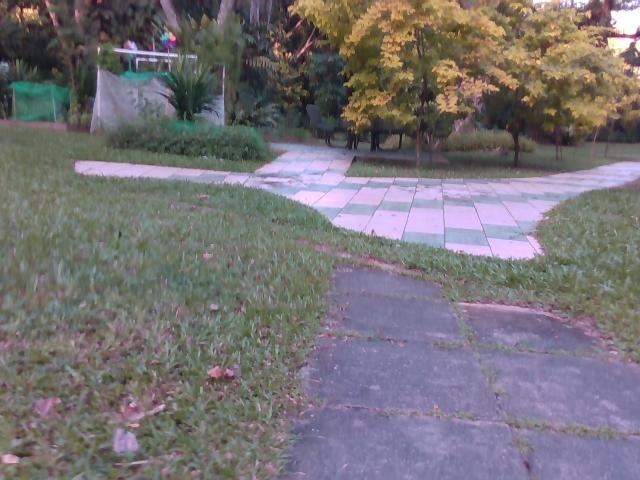}
    \includegraphics[width=0.19\linewidth]{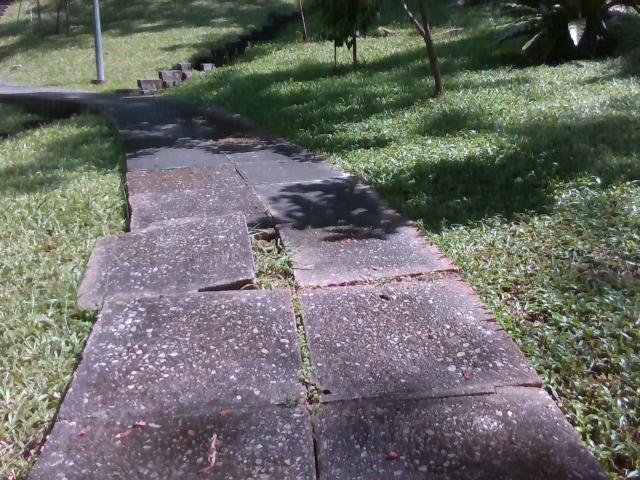} \\
  \end{tabular}
  \caption{The collected dataset spans a diverse array of indoor, semi-outdoor and outdoor environments. The diversity in this dataset enables the learned policy to potentially generalize to novel scenes. The dataset also contains challenging specular objects like glass doors and windows, and since it is collected with commodity Realsense cameras with limited dynamic range, also includes realistic sensor artifacts like over- and under-exposure of the image in certain areas.}
  \label{fig:data_collection_env}
\end{figure*}

The DECISION controller is learned by imitating expert intentions on trajectories collected by teleoperating the robot in the real world. We train it with the $L2$ loss over a human demonstration dataset $\mathcal{D} = \{\langle o_t, m_t, a_t \rangle\}_t$ comprising tuples of observations $o_t$, intentions $m_t$ and actions $a_t$ which is a tuple of linear and angular velocity. Specifically, we optimise
$$
    \mathcal{L} = \mathbb{E}_{\langle o_t, m_t, a_t \rangle \in \mathcal{D}} \|\pi_{m_t, c_{t}}(o_t) - a_t \|_2
$$

The dataset comprises RGB images collected from 3 RGB-D cameras mounted on our robot, together with the target intention and linear/angular velocities commanded by the human expert. We continuously expanded the dataset over 17 iterations in a DAgger-like manner~\citep{dagger}. Specifically we alternate between training and testing, iteratively collecting more data to cover both 1) observed failure modes and 2) new environments we are not able to generalise to. Our final dataset contains 727K time steps, or 2.2 million images, covering a wide range of indoor and outdoor environments and multiple terrains (\autoref{fig:data_collection_env}). Further details on the model training process can be found in \autoref{decision:training}.

\section{High-level planner and map system}
\label{sec:5}
In this section, we present the architecture of \intentionnet{}'s high-level planner and map system, which generate the intentions used to guide the low-level controller in a navigation task. The overall planner and map system are designed specifically to enable scalable planning and navigation to distant goals, by using compositionally generalisable classical planning algorithms for pathfinding over a `map-lite' environment representation.



\subsection{``Map-lite'' environment representation}
\begin{figure*}[t]
  \centering
\includegraphics[width=0.9\linewidth]{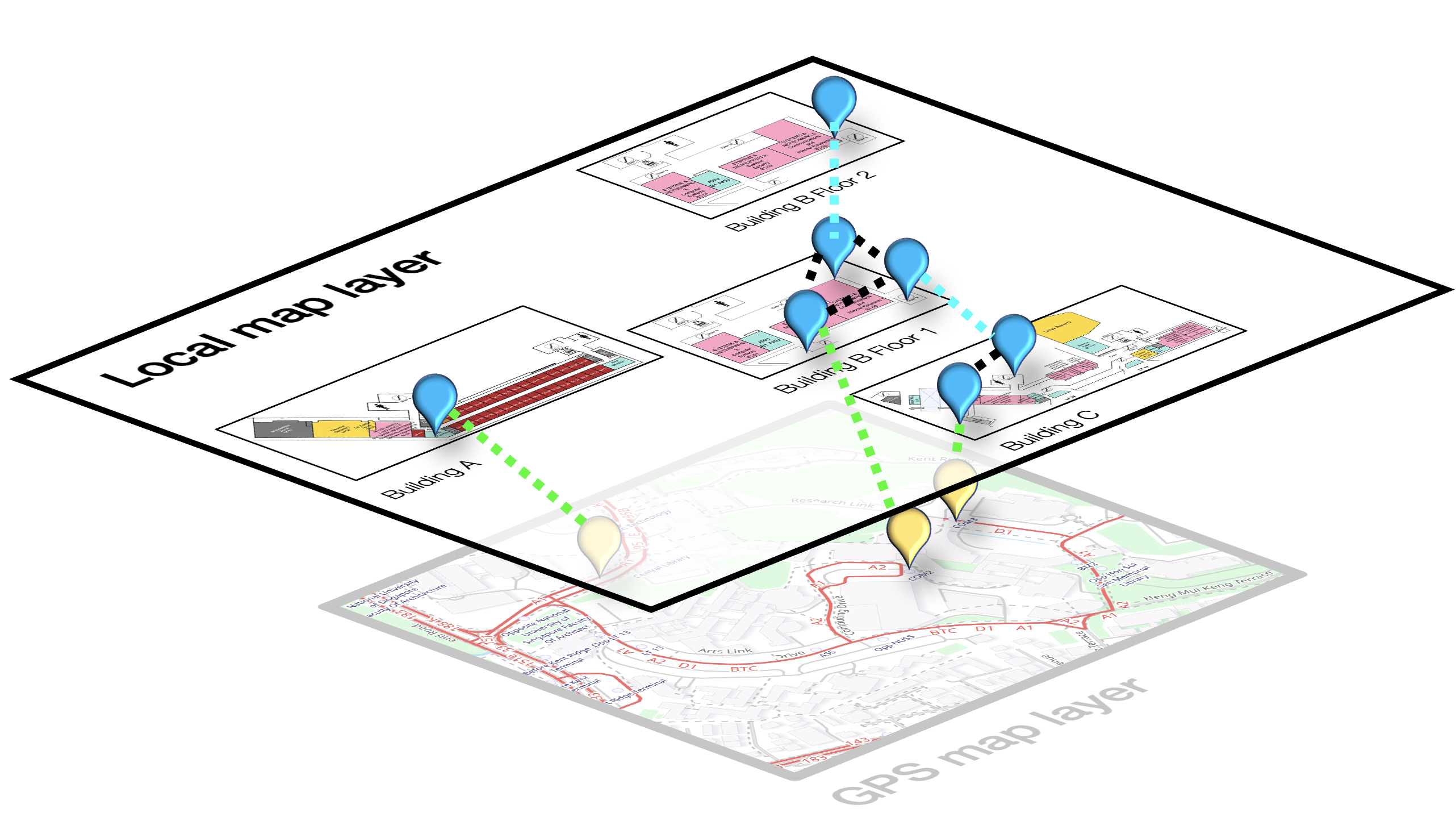}
  \caption{Example of our two-layer map system for handling indoor/outdoor map information. The GPS map layer contains a geo-referenced road network. The local map layer contains occupancy grids (derived from floor plans) of areas not detailed in the GPS map layer. Nodes are placed at the entrances/exits of each local map area, and a connectivity graph defined over these nodes shows how to traverse between these maps, and also grounds the local maps with respect to the GPS map layer.}
  \label{fig:map_representation}
\end{figure*}

\intentionnet{}'s map representation is ``map-lite'' in the sense that it can be constructed from various coarse and metrically inaccurate prior maps of the environment. This allows us to use readily available resources like floor-plans which are often passed over due to their coarseness and inaccuracy, and enables scalable map construction of large-scale areas.

The ``map-lite'' representation comprises two levels. At the top level, it contains a topological graph of paths spanning the entire environment that can be constructed from some global coarse map of the environment. At the lower level, it contains a collection of local coarse maps that each provide finer-grained information about paths in specific areas not represented in the global map. In the context of long-range navigation, the robot often is required to traverse both outdoor and indoor environments. The ``map-lite'' representation can capture such heterogeneous environments by representing the outdoor environment as a graph of paths extracted from some global map like road networks or satellite maps, and the indoor environments using local coarse maps like floor-plans or inaccurate and outdated occupancy grids.

\autoref{fig:map_representation} presents the specific instantiation of this ``map-lite'' representation employed in \inetlr{}. It consists of 3 components: (1) a GPS map layer, (2) a local map layer and (3) a connectivity graph defined over both the local and GPS map layers. The \textit{GPS map layer} contains a geo-referenced topological graph of outdoor walking paths extracted from OpenStreetMap road networks~\citep{osm}. The \textit{local map layer} encodes reachability information for specific areas for which the GPS map layer does not contain any information, such as the insides of buildings. The \textit{connectivity graph} represents connections between locations within the local coarse maps, and also connects locations in the local map layer to locations in the GPS map layer. We elaborate on the local map layer and connectivity graph below; further in-depth details about each layer and its method of construction can be found in \autoref{appendix:mapbuilding}.

The \textit{local map layer} is a collection of floor-plans, each representing a specific area such as a particular floor of a building. Floor-plans are used as they are readily available sources of prior information about indoor environments. While they are metrically inaccurate and hence unsuitable for use by many existing robot systems, the DLM-based \inetlr{} system can handle such inaccuracies, allowing for their use in our map representation. We analyse the metric inaccuracies that our system can tolerate in greater depth in \autoref{sec:system_robustness}. In \inetlr{}, we only process the floor-plans by binarizing them before storing them in the local map layer, which allows our planner to treat them directly as occupancy grids.


The \textit{connectivity graph} joins together the disparate pieces of map information, namely the various local coarse maps and the GPS map layer's graph, in a single unified representation. To plan paths beyond the localised area described by each local coarse map, we identify the \textit{Exits} in each local map from which we can transit to another local map, or to a path in the GPS map layer's graph. The connectivity graph is then an undirected graph where the nodes represent said \textit{Exits} and the edges connect the \textit{Exits} together. 

In particular, the \inetlr{} has 3 types of \textit{Exit} nodes: \texttt{stairs} - which allow for transits between different floors, \texttt{linkways} - which allow for transit between different buildings, and \texttt{outdoors} - which denotes a transit to a path in the GPS map layer's graph. Empirically, we find these types of \textit{Exits} to be sufficient to describe the varied environments we test our system in. \inetlr{} also employs 3 types of edges, highlighted in \autoref{fig:map_representation}. Black edges represent intra-map edges, which keep track of the travel distance between \textit{Exits} in the same local coarse map with their edge weights. When initialising the map representation, all \textit{Exit} nodes within the floor-plan are connected together in a complete graph with intra-map edges. The distance edge weights between the nodes are updated online during navigation with odometry, to provide more accurate distance estimates than the floor-plan itself provides. Cyan edges are inter-map edges that connect \textit{Exit} nodes on different local coarse maps (either \texttt{stairs} or \texttt{linkways} nodes). Finally, green edges are inter-layer edges that connect an \texttt{outdoors} \textit{Exit} node in a floor-plan to the corresponding \textit{Exit} node on the GPS map layer's graph.

\subsection{Planning and intention generation}
\label{sec:planning_intgen}

\intentionnet{} employs a hierarchical planning approach using classical planning algorithms to enable compositional generalisation and scalability of planning to distant goals. In \intentionnet{} systems, goals may be specified either as a GPS location on the GPS map layer, or as a point on a local coarse map in the local map layer. The first step in planning is to search over the GPS map layer's graph and the connectivity graph to obtain a rough topological path. If the robot's current position or goal is on a local coarse map, we run A$^*$ over the local map to obtain a path to the nearest reachable \textit{Exit} node first, before computing the rough topological path from the nearest \textit{Exit} node. Once we have a rough topological path, we convert each edge on the path into intentions. This intention ``plan'' is then used to guide the low-level controller during navigation. In particular, the system uses off-the-shelf odometry (\autoref{sec:5_loc}) to track its metric position and issue the right intention to the low-level controller at the appropriate time. 

\begin{figure}[t]
  \centering
    \includegraphics[width=0.99\linewidth]{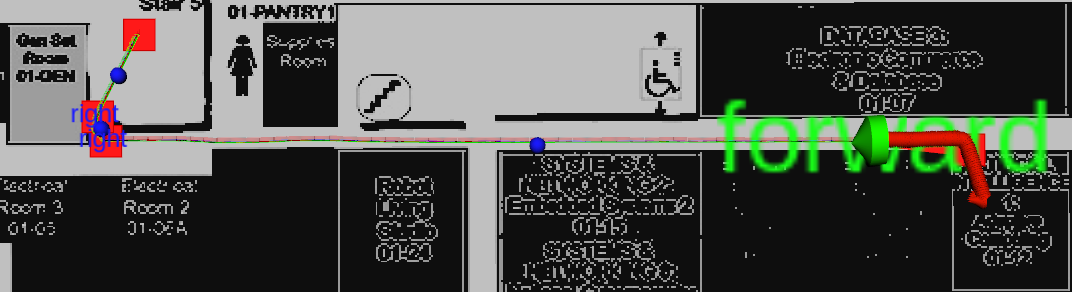}
    \caption{Intention generator example on one sample \textit{local map layer}. The green arrow is the current robot position. The purple line is the planned path from \textit{Exit} to another \textit{Exit}. The red rectangles are the control points of Ramer–Douglas–Peucker algorithm (RDP), and the aligned text are the intentions around that control point. The blue points are the middle of two control points. In our experiments, we set the radius of the intention influence to be a minimum of 5 meters and the distance of the nearest midpoint.}
  \label{fig:intention_gen}
\end{figure}

We specifically describe the approach taken in the \inetlr{} system to convert the rough topological path from the planner into an intention ``plan''. DLM intentions can be considered as a discretisation of a path's curvature into 3 coarse bins corresponding to \{\texttt{left}, \texttt{forward}, \texttt{right}\}. We use the angle subtended between each consecutive pair of line segments along a path as a proxy metric for the curvature, and threshold on the angle to assign the appropriate DLM intention. Paths between two points on the GPS map layer are already represented as polylines, so we can directly apply the approach above to obtain the intentions along such paths. For intra-map edges or edges between points on the same local coarse map, we first use A$^*$ on the binarised floor-plan to find the path between the nodes. As shown in \autoref{fig:intention_gen}, the A$^*$ path can then be converted into a coarser polyline using Ramer-Douglas-Peucker (RDP) polyline simplification~\citep{ramer_rdp, douglas_rdp}, allowing the intention generation approach above to be applied. As inter-map and inter-layer edges ideally connect \textit{Exit} nodes that occupy the same spatial position, we do not generate any intentions from them.



\subsection{Localisation}
\label{sec:5_loc}
Our system maintains two modes of localization: GPS and odometry. When navigating outdoors, the robot uses GPS to track its position in UTM coordinates with respect to the geo-referenced GPS map layer graph. When traversing paths near or inside buildings, where GPS signals may not be present or may provide inaccurate readings due to effects such as multipath, the robot relies instead on odometry. Specifically, we use the Elbrus visual-inertial odometry (VIO) module from Nvidia's ISAAC SDK~\citep{elbrus} to continuously track the robot's 6-DoF pose using observations from our onboard stereo camera and IMU. We note that while Elbrus has a loop closure component that enables a complete SLAM system with relocalisation, we only use the VIO module. This is because our system does not assume the environment is pre-mapped with existing SLAM techniques and therefore does not assume the availability of landmarks for relocalisation. As such, our robot navigates with continuously drifting odometry when indoors or near buildings.

\section{Experiments}

We explore the following 3 questions:
\begin{enumerate}[label=\textbf{Q\arabic*},ref=\textbf{Q\arabic*}]
    \item \label{qn:robustness} Does learning a low-level controller enable more \textit{robust control and obstacle avoidance} than classical low-level controllers?
    \item \label{qn:intentions} What \textit{capabilities} do different intention designs enable?
    \item \label{qn:inaccuracies} How do \intentionnet{} and classical architecture types compare in terms of \textit{scalability to long-range navigation}?
\end{enumerate}


In general, we find that our learned low-level controllers are capable of robust obstacle avoidance and are better able to model the robot's dynamics than classical controllers. We further verify the effectiveness of a DLM-based controller in tolerating metric inaccuracies, and show that an \intentionnet{} system built on a DLM controller can successfully navigate even in the presence of significant inaccuracies in mapping and localisation.

\subsection{Experimental setup}
\begin{figure}[t]
  \centering
\includegraphics[width=0.98\linewidth]{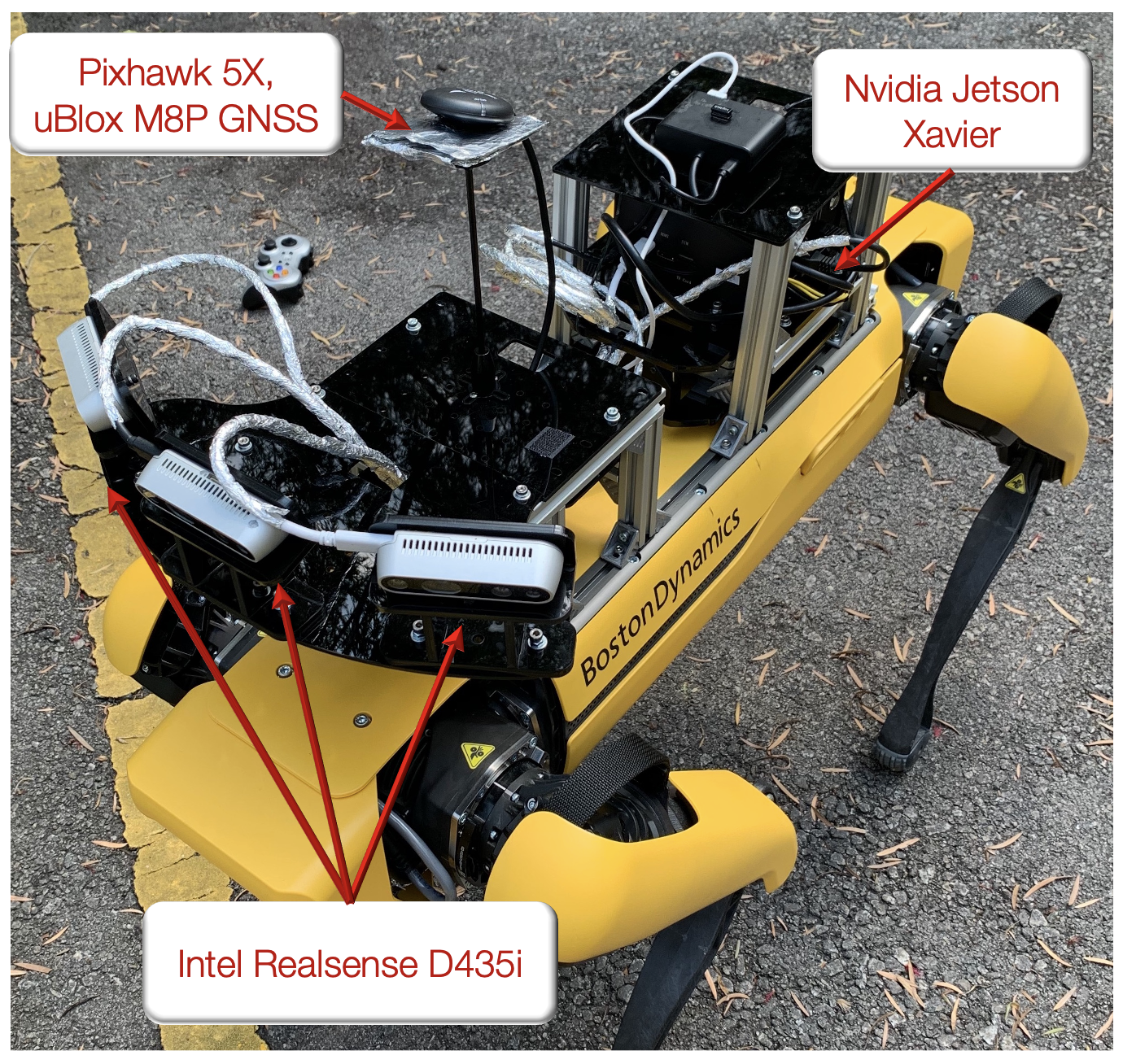}
  \caption{Boston Dynamics Spot robot hardware setup}
  \label{fig:spot_hardware}
\end{figure}

For the experiments in simulation, we use a simulated Turtlebot equipped with a 70$^{\circ}$ field-of-view webcam and a Sick LiDAR (used purely for localization), and run the tests in the Stage simulator~\citep{vaughan_stage}.

For the real-world tests we use a Boston Dynamics Spot robot, for its ability to traverse many challenging types of terrain and climb stairs. Our hardware setup is shown in \autoref{fig:spot_hardware}. For sensing, we use 3 Intel Realsense D435i RGB-D cameras, which provide a 140$^{\circ}$ horizontal field-of-view of the robot's front. Our navigation system runs completely on an Nvidia AGX Xavier embedded computer mounted on the Spot robot. For navigation tasks that require GPS, we use a mounted Pixhawk 5X with a uBlox M8P GNSS unit.

The Spot robot provides innate collision avoidance capabilities that activates at very close range to obstacles ($< 10$cm). As an additional safety layer, we keep the Spot's collision avoidance on throughout our experiments. Our DECISION low-level controller is usually capable of avoiding at distances much earlier than 10cm, and we count episodes where the Spot gets to within 10cm of an obstacle as failures.

We evaluate the performance of our system using the following metrics: 1) Success Rate (SR) which is the percentage of successful trials, 2) Average Intervention Count (Avg. Int.) which is the average number of human interventions required per trial, 3) Average Execution Time (Avg. Ex. Time) which is the average time required to complete a trial, and 4) Smoothness which is the average jerk of the trajectories traversed by the robot.

\subsection{Robustness of low-level controller}
To answer \ref*{qn:robustness}, we compare the robustness of control and obstacle avoidance across learned and classical, handcrafted low-level controllers. In particular, we explore our hypotheses that 1) a learned controller is capable of more \textit{robust control} because it has a better robot dynamics model, and that 2) learned low-level controllers are capable of effectively capturing salient signals from complex, high-dimensional RGB inputs for \textit{robust obstacle avoidance}. We first establish the benefits of learning a controller by testing both learned and classical low-level controllers in simulation and the real world. We then stress test our learned controllers on even more challenging partially-observable real-world situations to establish their robustness in realistic scenarios.

\subsubsection{Simulation experiments}
\begin{figure*}
    \centering
    \subfloat[]{
        \includegraphics[width=0.18\textwidth]{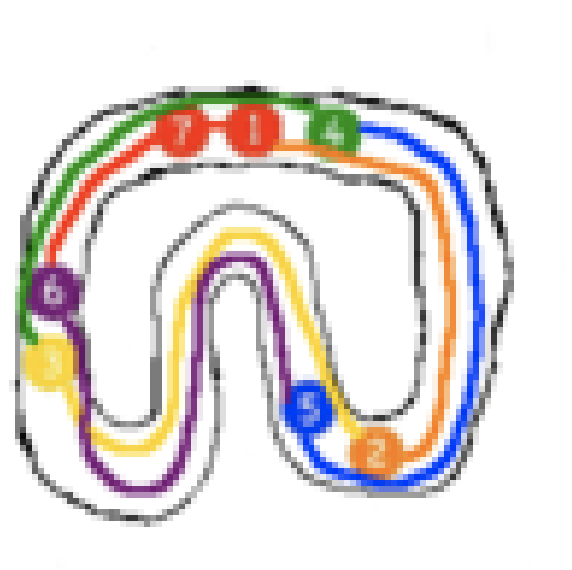}
    }\hfill
    \subfloat[]{
        \includegraphics[width=0.18\textwidth]{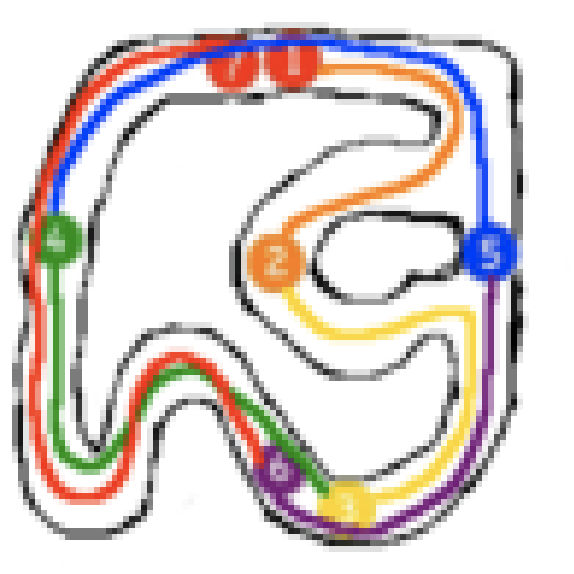}
    }\hfill
    \subfloat[]{
        \includegraphics[width=0.18\textwidth]{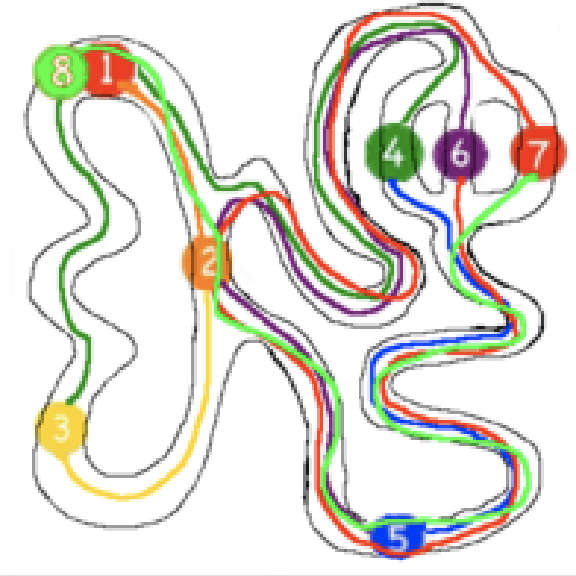}
    }\hfill
    \subfloat[]{
        \includegraphics[width=0.18\textwidth]{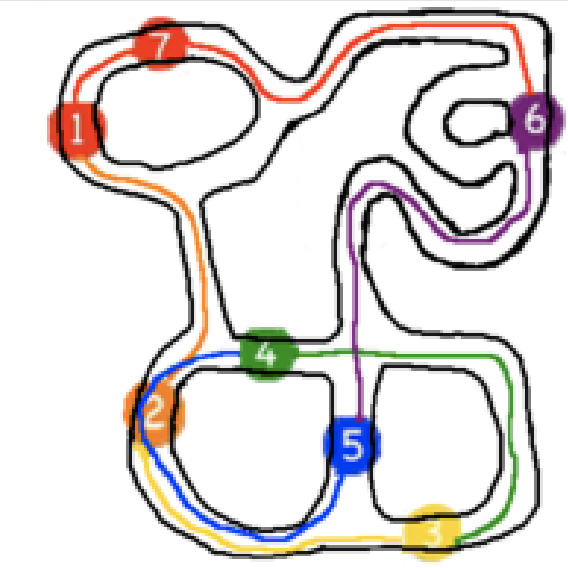}
    }\hfill
    \subfloat[]{
        \includegraphics[width=0.18\textwidth]{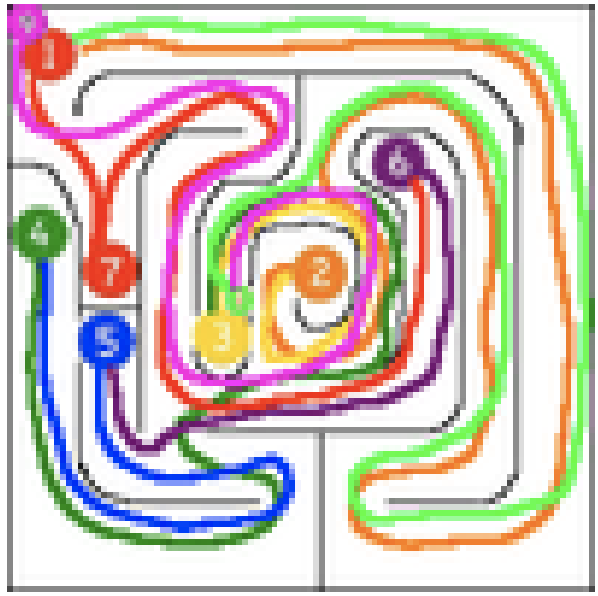}
    }
    \caption{Five different navigation tasks used to compare the performance of the DLM and LPE intention types in simulation. Each task requires the robot to navigate in sequence through the numbered waypoints. The planned path from which the intentions are generated is overlaid as coloured lines.}
    \label{fig:sim_controllers}
\end{figure*}

We compare our learned low-level controllers against two classical controller baselines. The learned controllers are the DLM- and LPE-steered controllers described in \autoref{sec:controller_steering}. The classical baselines are 1) a PID controller that tracks a planned path but has no reactive obstacle avoidance capabilities, and 2) a Dynamic Window Approach (DWA)~\citep{fox_dwa} controller that both tracks a planned path and uses instantaneous LiDAR scans for reactive obstacle avoidance. The PID and DWA controller's parameters are both tuned for the Turtlebot robot model used in simulation. We hand-sketched 5 different maps and generated 3D walled environments in simulation. The controllers were tasked with following a specified path on each map, marked out in coloured lines on \autoref{fig:sim_controllers}. If the robot collides with the environment, we reset it to the nearest position on the specified path and record this as an intervention.

From the results in \autoref{tab:compare_controllers_sim}, we find that learned controllers outperform classical hand-tuned controllers with faster completion times and smoother trajectories on most paths tested. The model-based, hand-tuned PID and DWA controllers are only able to slightly outperform the learned controllers on Task (a) which comprises a few smooth, gentle curves and is less challenging than Tasks (b)-(e). This suggests that learned controllers are better able to capture the robot's dynamics, allowing them to generate actions that produce more efficient and smoother trajectories. We also find that DLM, LPE and DWA are all capable of safely navigating the robot without collisions, verifying our hypothesis that learned low-level controllers are able to learn to use rich RGB inputs for effective reactive obstacle avoidance.


\begin{table*}[t]
\small\sf\centering
\caption{Comparing robustness of control and obstacle avoidance across \textcolor{blue}{learned} and \textcolor{orange}{classical} low-level controllers}
  \label{tab:compare_controllers_sim}
 \begin{tabular}{l c c c c c c c c c c c c}
    \toprule
    Task &  
    \multicolumn{4}{c}{Interventions ($\downarrow$)} &  
    \multicolumn{4}{c}{Avg. Ex. Time/secs ($\downarrow$)} & 
    \multicolumn{4}{c}{Smoothness ($\downarrow$)} \\
    \cmidrule(lr){2-5} \cmidrule(lr){6-9} \cmidrule(lr){10-13}
    & \textcolor{blue}{DLM} & \textcolor{blue}{LPE} & \textcolor{orange}{PID} & \textcolor{orange}{DWA} & \textcolor{blue}{DLM} & \textcolor{blue}{LPE} & \textcolor{orange}{PID} & \textcolor{orange}{DWA} & \textcolor{blue}{DLM} & \textcolor{blue}{LPE} & \textcolor{orange}{PID} & \textcolor{orange}{DWA}\\
    \midrule
    \textit{(a)} & 0 & 0 & 0 & 0 & 113 & 114 & \textbf{105} & 109 & 0.0053 & 0.0053 & 0.0040 & \textbf{0.0039}\\
    \textit{(b)} & 0 & 0 & 3 & 0 & \textbf{118} & 128 & 155 & 126 & 0.0074 & \textbf{0.0068} & 0.0170 & 0.0100\\
    \textit{(c)} & 0 & 0 & 16 & 0 & \textbf{559} & 561 & 640 & 565 & 0.0074 & \textbf{0.0072} & 0.0150 & 0.0094\\
    \textit{(d)} & 0 & 0 & 5 & 0 & 237 & \textbf{226} & 240 & 238 & 0.0085 & \textbf{0.0066} & 0.0120 & 0.0095 \\
    \textit{(e)} & 0 & 0 & 21 & 0 & 545 & \textbf{531} & 546 & 551 & 0.0080 & \textbf{0.0075} & 0.0089 & 0.0084\\ 
    
 \bottomrule
\end{tabular}
\end{table*}

\subsubsection{Real-world experiments}
We compare a DLM-steered controller (DECISION) and DWA to validate our hypotheses in the real world. We also test DECISION in several challenging and realistic scenarios to demonstrate that it can learn control and obstacle avoidance capabilities robust enough for practical deployment in the real world.

\textbf{Comparing DECISION and DWA.} The \textit{Original} setting in \autoref{tab:nav_robustness} tests the robustness of DECISION (abbreviated DEC) against DWA on goal-reaching in a cluttered office environment. Further details of the experimental setup are given in \autoref{sec:system_robustness}. We observe that DECISION outperforms DWA with faster execution times, smoother trajectories and fewer interventions. DECISION's better performance in terms of speed and smoothness suggest that it has learned to capture a better model of real-world robot dynamics than DWA. With fewer interventions needed, DECISION is not only effective at learning to use complex RGB inputs for obstacle avoidance, but is more effective at it than classical controllers like DWA.

\begin{figure*}[t]
  \centering
  \setlength{\tabcolsep}{0.5pt} 
  \renewcommand{\arraystretch}{1} 
  \begin{tabular}{ccccccc}
    (\subfig{a})
    \raisebox{-.5\height}{\includegraphics[width=0.158\linewidth]{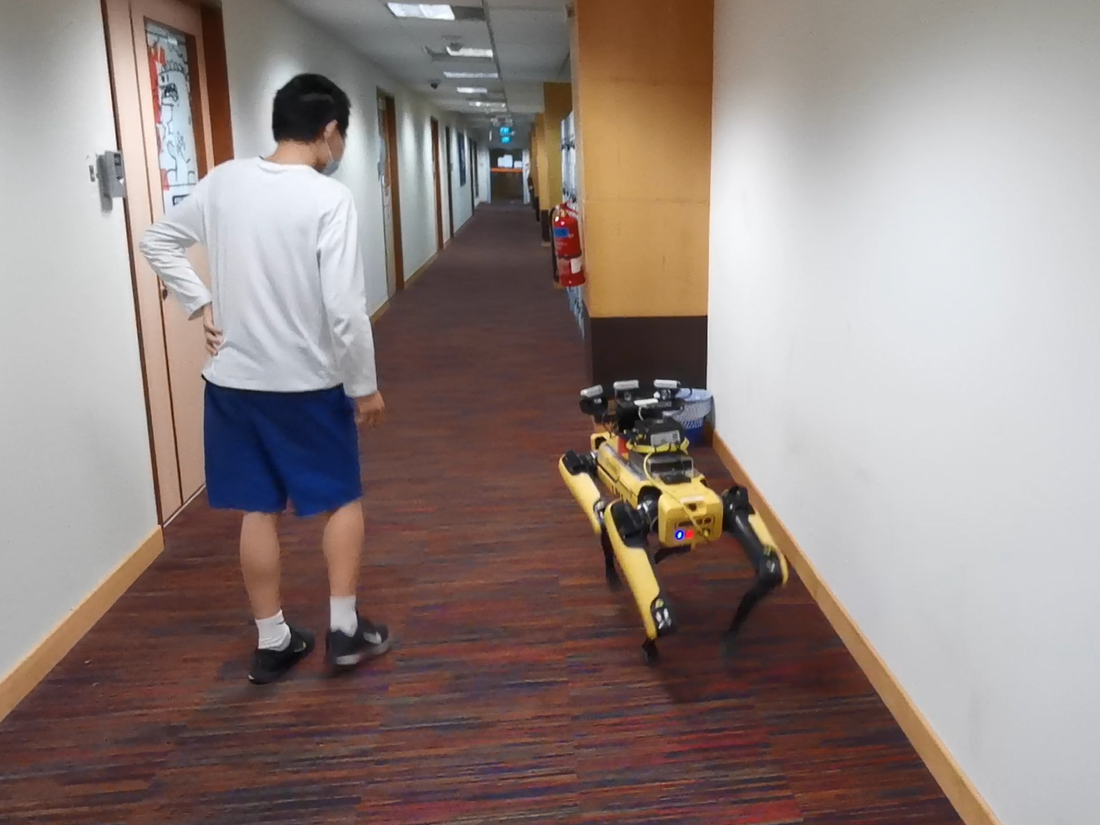}} 
    \raisebox{-.5\height}{\includegraphics[width=0.158\linewidth]{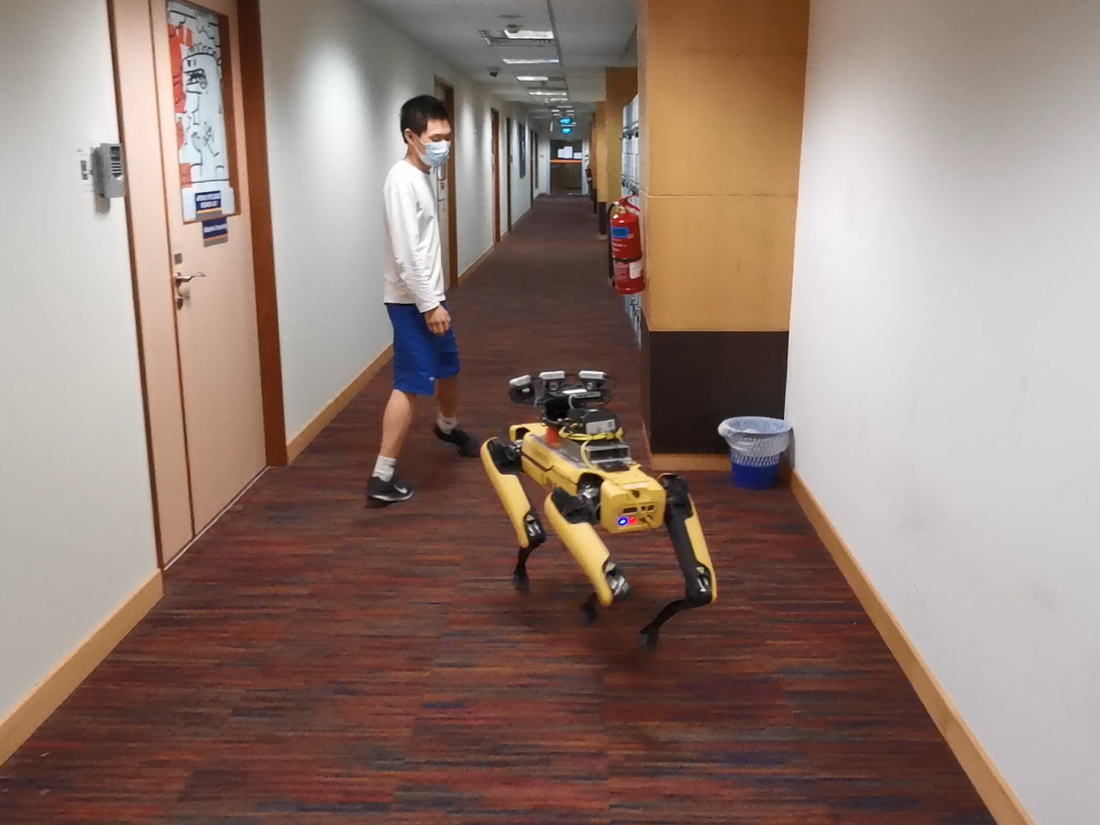}} 
    \raisebox{-.5\height}{\includegraphics[width=0.158\linewidth]{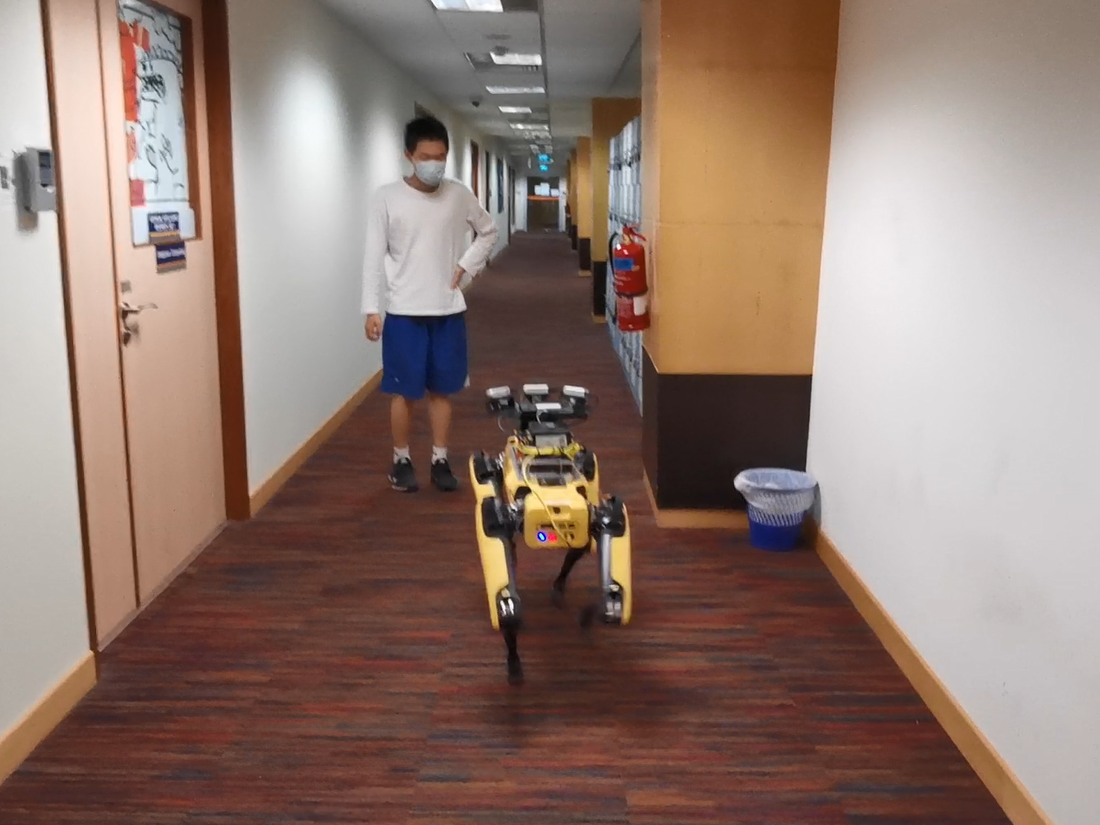}} 
    \raisebox{-.5\height}{\includegraphics[width=0.158\linewidth]{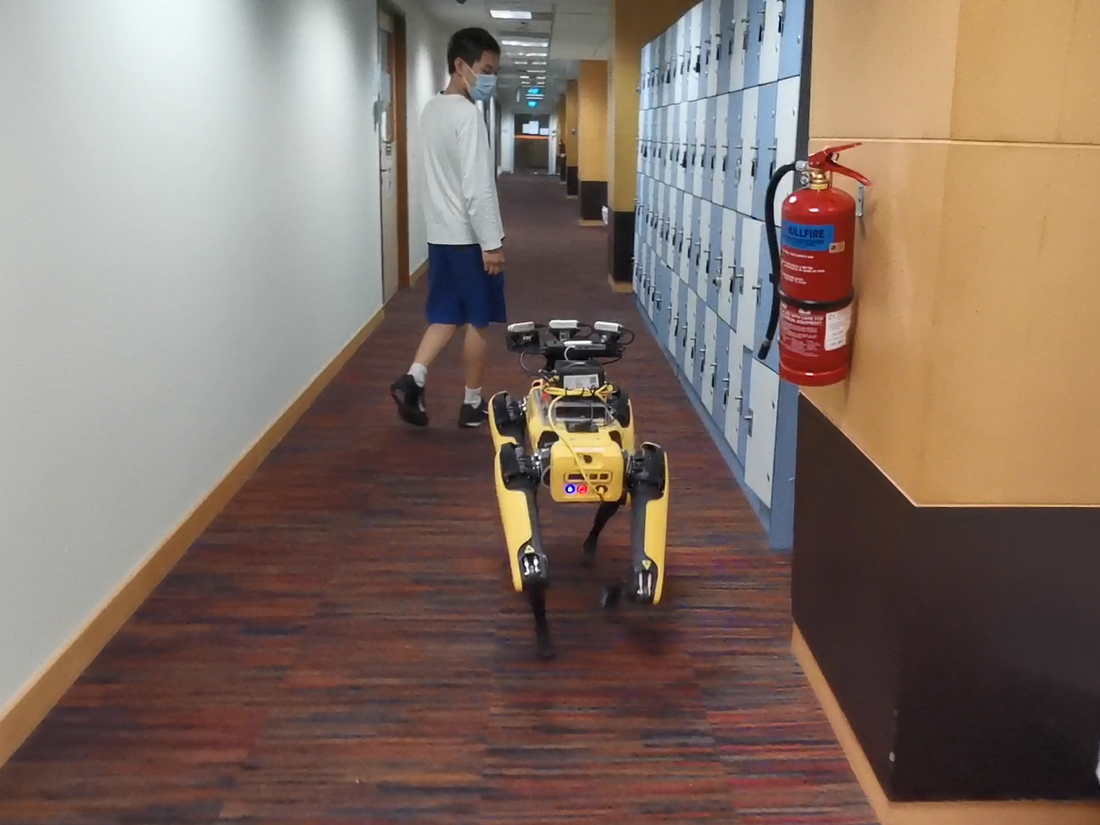}} 
    \raisebox{-.5\height}{\includegraphics[width=0.158\linewidth]{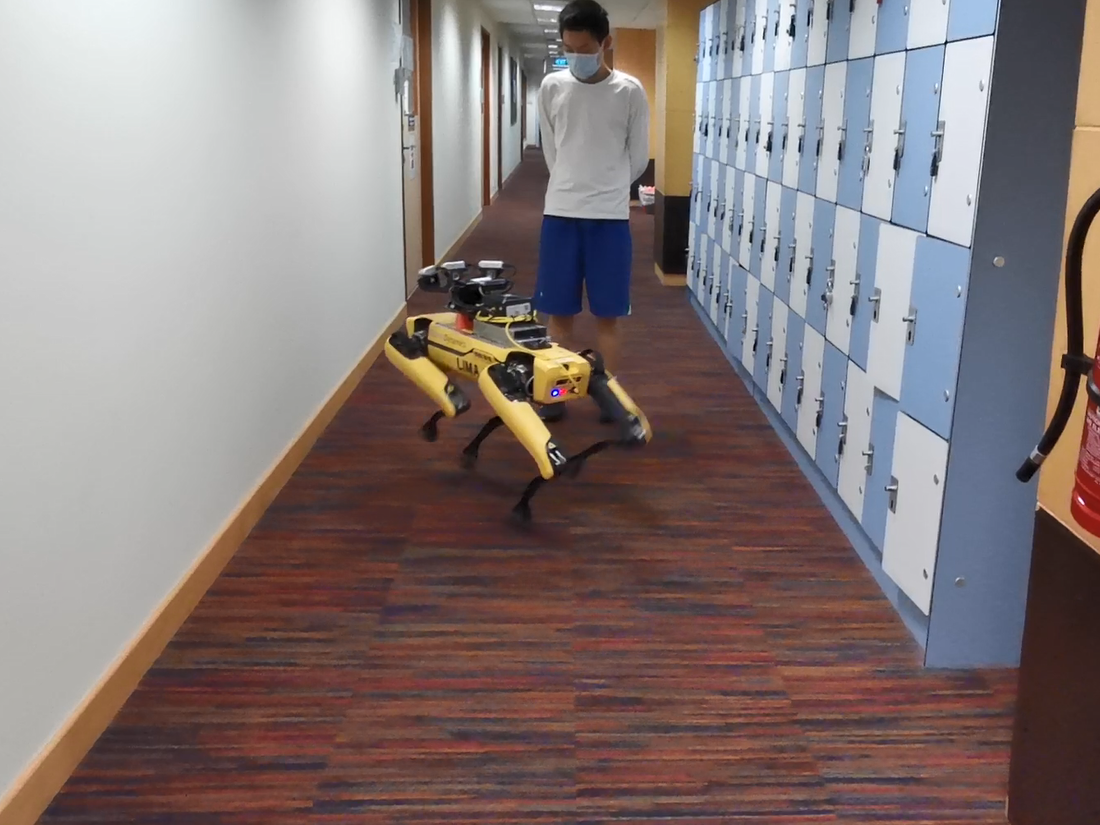}} 
    \raisebox{-.5\height}{\includegraphics[width=0.158\linewidth]{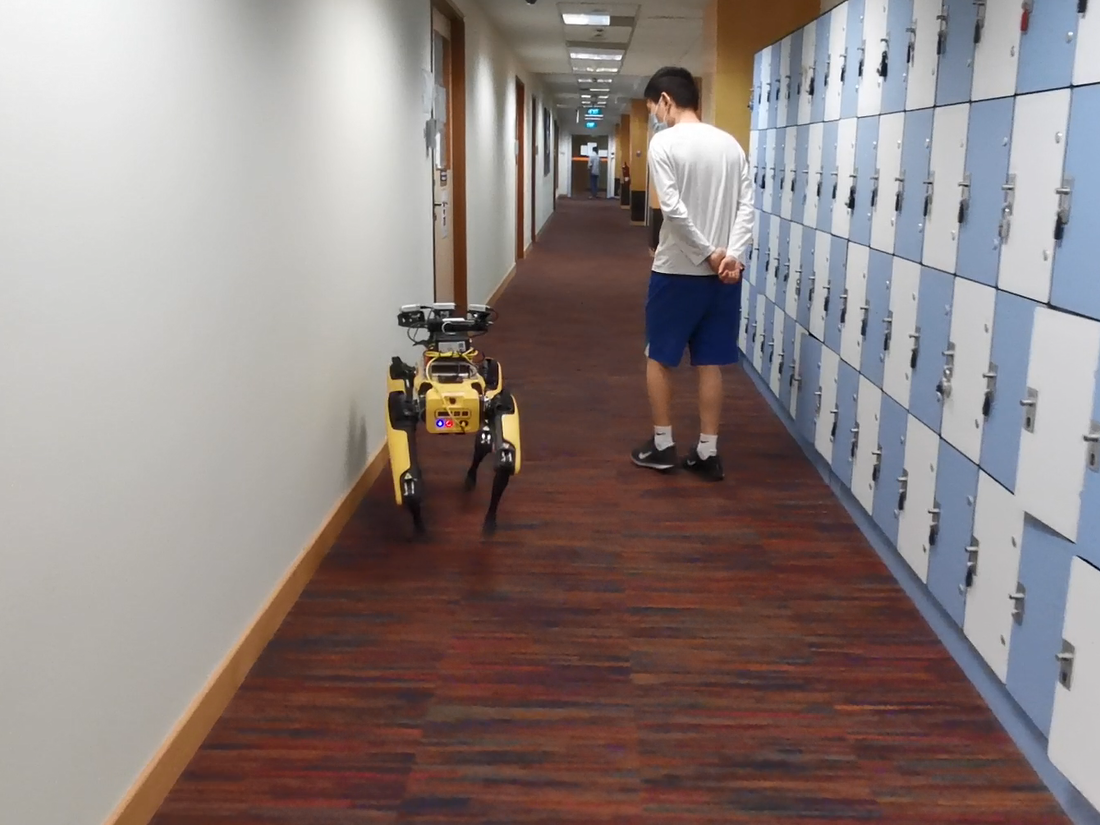}} \vspace{2pt} \cr
    (\subfig{b}) 
    \raisebox{-.5\height}{\includegraphics[width=0.158\linewidth]{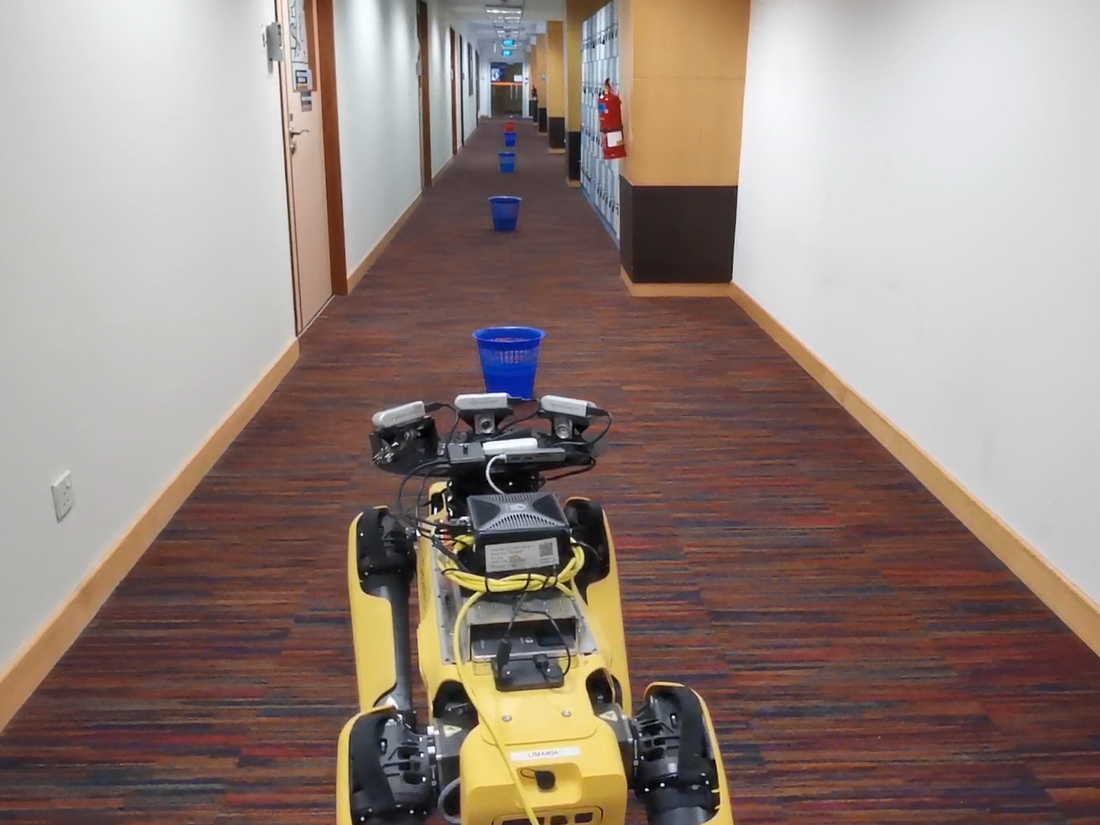}} 
    \raisebox{-.5\height}{\includegraphics[width=0.158\linewidth]{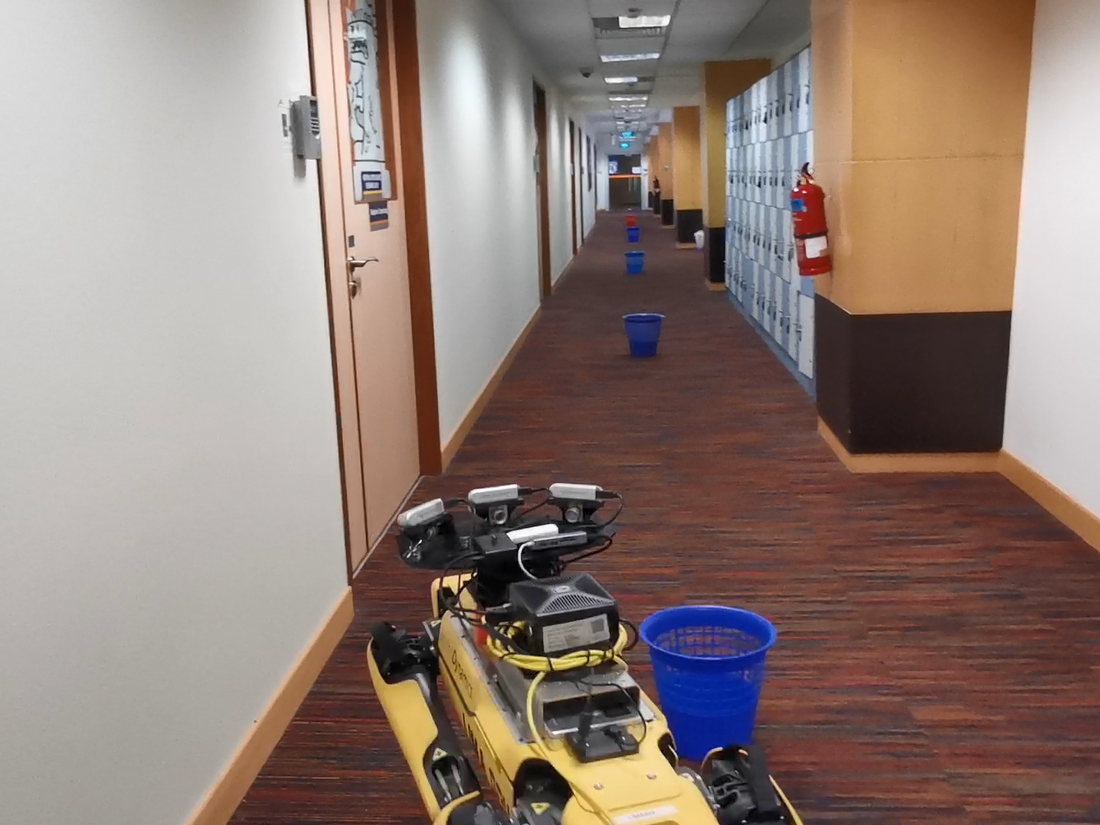}} 
    \raisebox{-.5\height}{\includegraphics[width=0.158\linewidth]{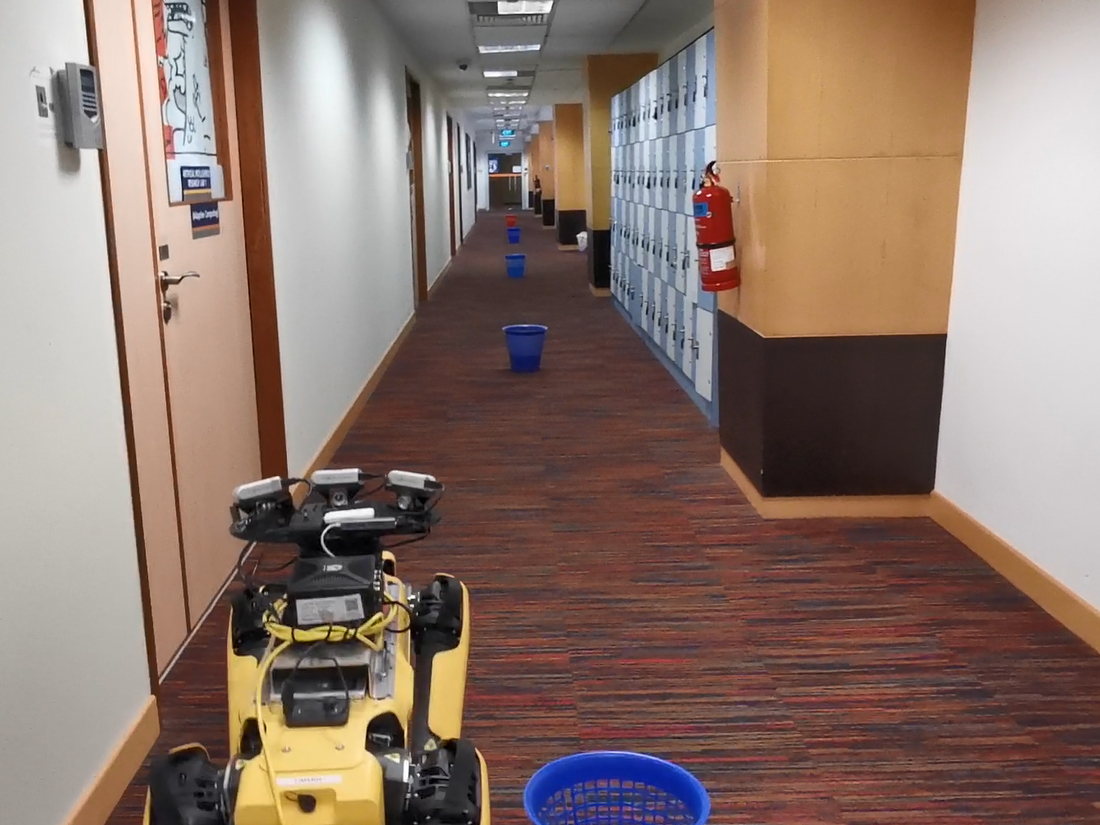}} 
    \raisebox{-.5\height}{\includegraphics[width=0.158\linewidth]{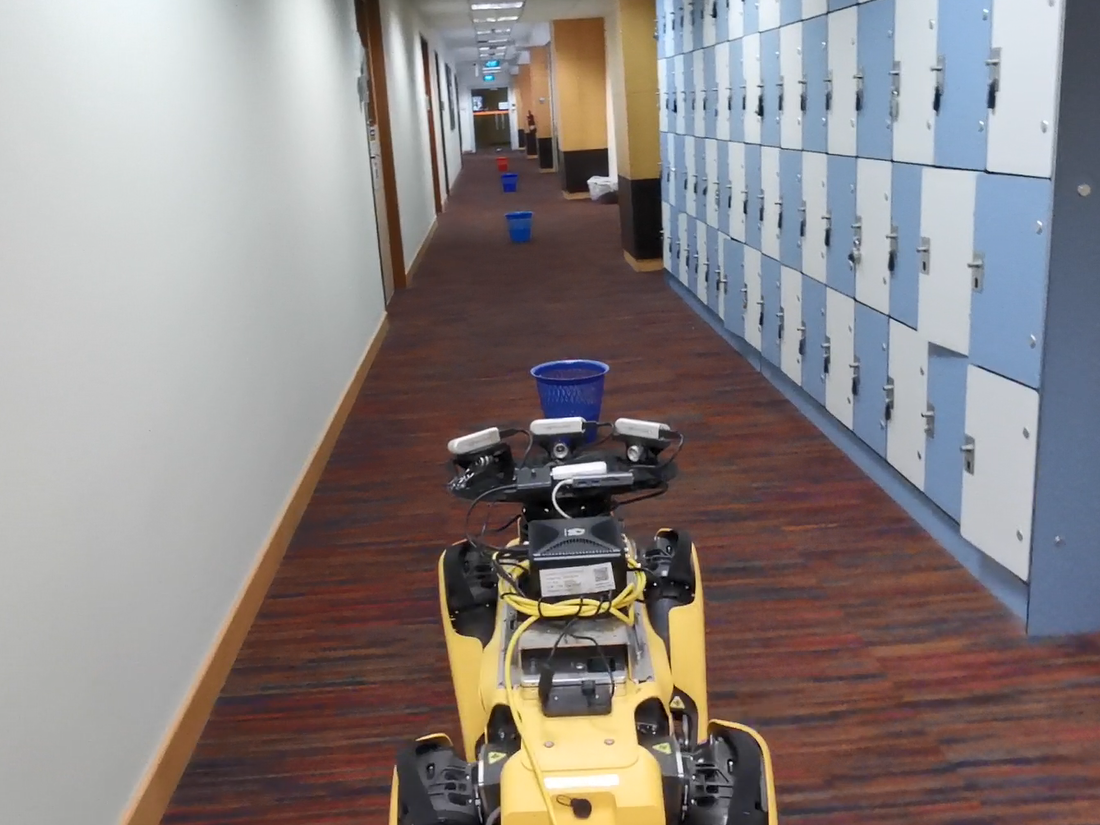}}
    \raisebox{-.5\height}{\includegraphics[width=0.158\linewidth]{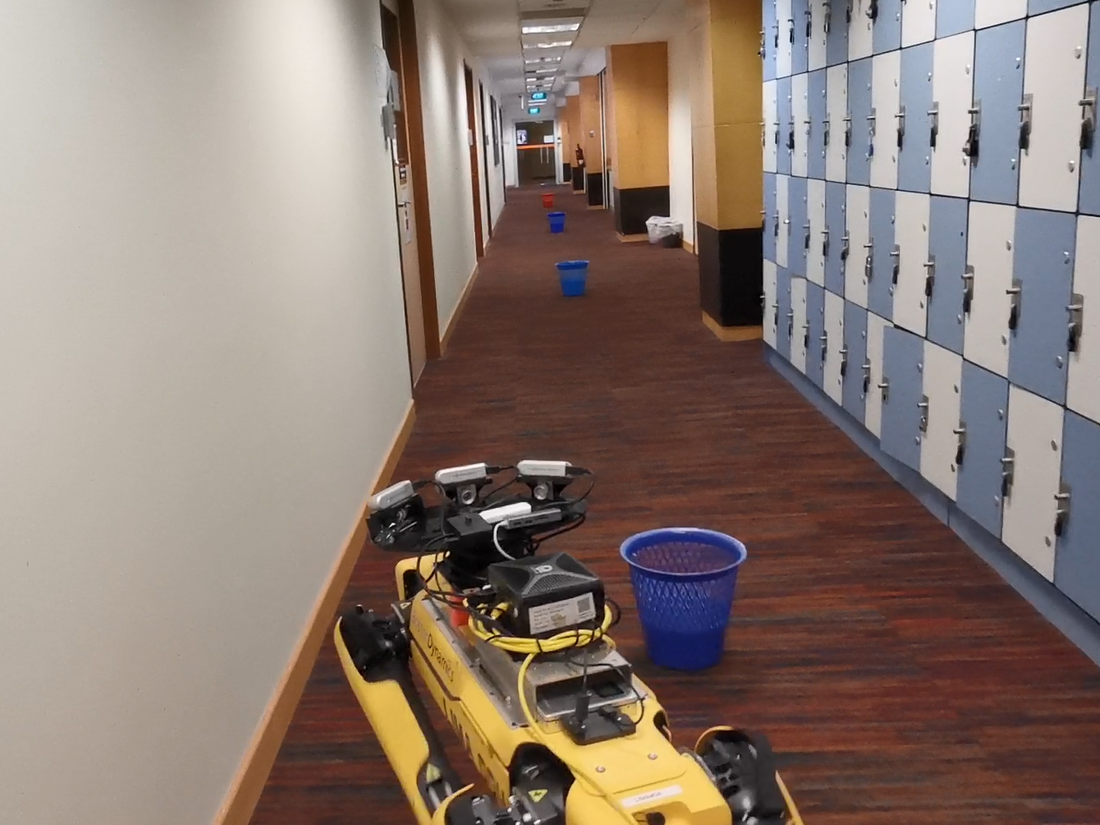}} 
    \raisebox{-.5\height}{\includegraphics[width=0.158\linewidth]{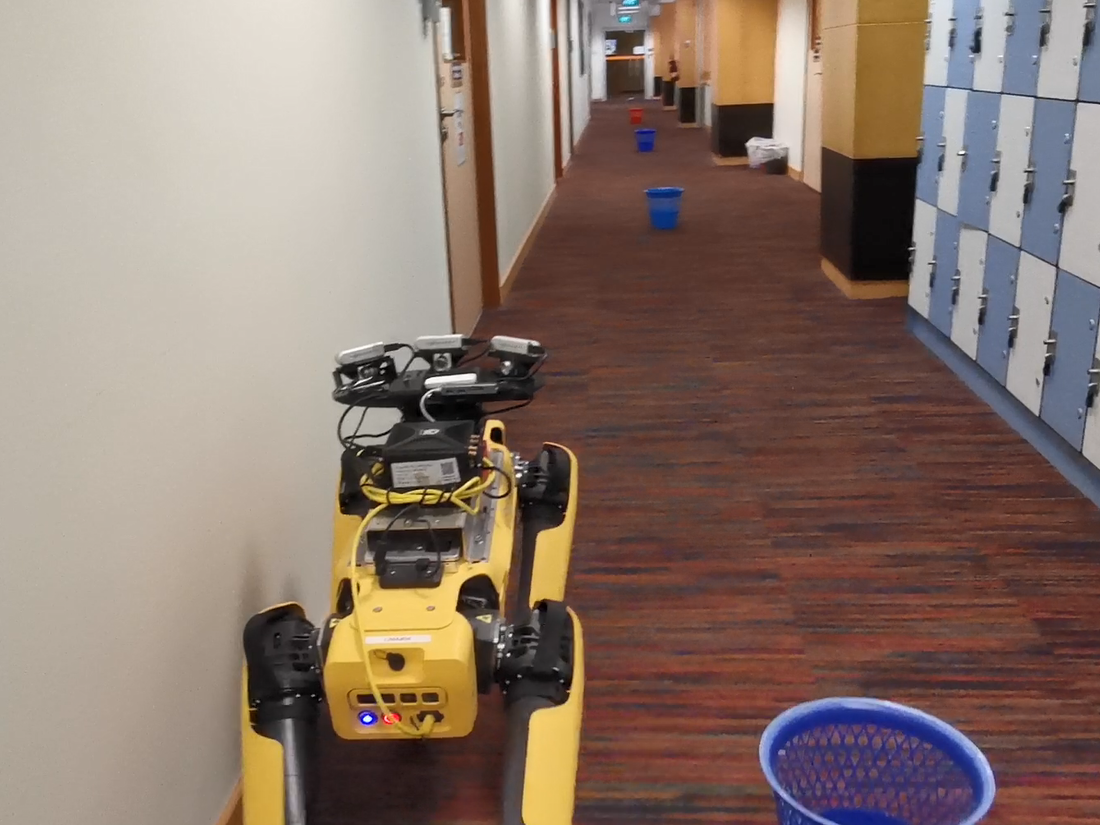}}  \vspace{2pt} \cr
  \end{tabular} 
  \caption{Illustrations of challenging real-world scenarios. (\subfig{a}) \taskref{pedestrian}: The robot avoids the adversary by predicting its movement. (\subfig{b}) \taskref{trashbin}: The robot successfully bypasses the basket that is not in the robot camera view. }
\end{figure*}

\textbf{Challenging scenarios for obstacle avoidance.} We test the DECISION controller on 2 realistic and challenging scenarios. \autoref{fig:controller_realworld} illustrates the 2 chosen scenarios. The details are as follows:
\begin{enumerate}[label=\roman*.]
    \item \textit{Adversarial pedestrian avoidance.} \label{pedestrian} This tests the DLM-steered controller's robustness at avoiding dynamic obstacles. In this scenario, the robot has to navigate down a 2.5m wide corridor, while avoiding collisions with an adversarial pedestrian who will attempt to block its path multiple times. Each time, the pedestrian will walk directly into the path of the robot, around 1-1.2m ahead of it, and success is recorded if the robot is able to adjust its heading and avoid colliding with the pedestrian. Otherwise, the human operator will intervene to reorient the robot and continue with the test. In total, the pedestrian will block will attempt to block the robot around $n\approx 15$ times in a test. We report the SR, i.e. the percentage of pedestrian blockages the robot successfully navigates around.
    
    \item \textit{Blind-spot object avoidance.} \label{trashbin} 
    This tests the DLM-steered controllers' robustness at avoiding static obstacles when partial observability plays a significant role. Specifically we test the controller's ability to avoid small objects on the ground much lower than the robot's camera height, such as a 20cm tall wastepaper basket. Due to the sensors' limited field-of-view, these objects will be in the robot's blind spot as it comes close to them ($\sim$60cm away), requiring that it use history information to avoid them. During this scenario, we line up $n=5$ wastepaper baskets at 5m intervals in a 2.5m wide corridor. A success is counted for each basket that the robot avoids; otherwise the human operator intervenes to navigate the robot around the basket. We report the SR, i.e. the percentage of baskets the robot successfully navigates around.
    
\end{enumerate}

\autoref{tab:controller_realworld} details the results of these tests. The DECISION controller attains a success rate and average intervention count on par with that of a human directly teleoperating the robot on both tasks. This suggests that the DECISION controller is capable of control and obstacle avoidance under partial observability at close-to-human levels of performance. It is robust in realistic, challenging and partially observable scenarios and can be practically deployed in real world settings.


\begin{table}[hb]
\small\sf\centering
\caption{Performance of low-level controllers on challenging real-world scenarios, over $N=10$ trials.}
\label{tab:controller_realworld}
 \begin{tabular}{l c c c c c}
    \toprule
    Model &  
    \multicolumn{2}{c}{\taskref{pedestrian}} & 
    \multicolumn{2}{c}{\taskref{trashbin}} \\
    \cmidrule(lr){2-3} \cmidrule(lr){4-5}
      & \SR{}($\uparrow$) & \CR{}($\downarrow$) & \SR{}($\uparrow$) & \CR{}($\downarrow$) \\ \midrule
    \textbf{\proposed{}} & \textbf{90} & \textbf{0.2} & \textbf{90} & \textbf{0.1} \\
    \human{}      & 90 & 0.2 & 90 & 0.1\\ 
 \bottomrule
\end{tabular}
\end{table}


\subsection{Capabilities induced by different intentions}
\label{sec:compare_intention}
We draw on the simulation results from \autoref{tab:compare_controllers_sim} to answer \ref*{qn:intentions}. We observe that both DLM- and LPE-steered controllers are comparable in terms of Average Execution Time and hence efficiency. However, LPE-steered controllers consistently execute smoother trajectories than DLM-steered controllers. This is likely because the LPE intentions provide not only environmental information, but also information on path history and desired future path. Compared to DLM, the richer information in LPE likely provides the extra context and guidance that enables the controller to navigate the robot more smoothly and precisely.


\subsection{Tolerance of mapping and positioning inaccuracies}
\label{sec:system_robustness}

\begin{figure*}[ht!]
    \centering
    \subfloat[Original]{
        \includegraphics[width=0.31\textwidth]{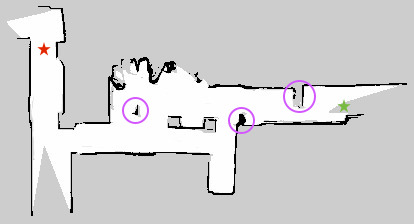}
        \label{subfig:original}
    }\hfill
    \subfloat[No obstacles + Scaled to 1.5 AR]{
        \includegraphics[width=0.31\textwidth]{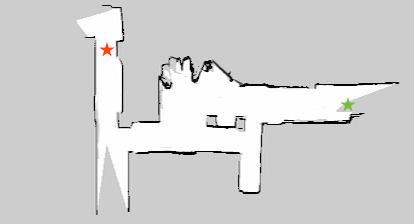}
        \label{subfig:scaled_1.5}
    }\hfill
    \subfloat[No obstacles + Hand-drawn]{
        \includegraphics[width=0.31\textwidth]{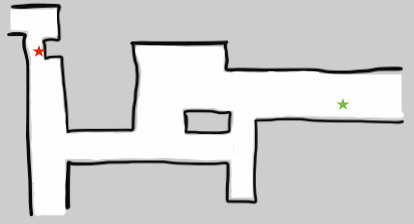}
        \label{subfig:hand_drawn}
    } \\

    \subfloat[No obstacles]{
        \includegraphics[width=0.31\textwidth]{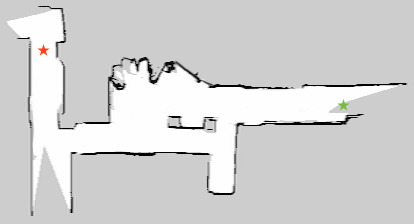}
        \label{subfig:no_obs}
    }\hfill
    \subfloat[No obstacles + Scaled to 1.33 AR]{
        \includegraphics[width=0.31\textwidth]{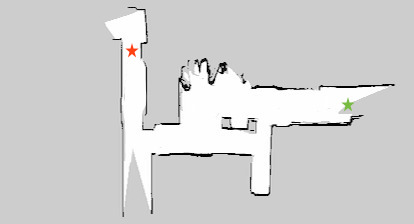}
        \label{subfig:scaled_1.33}
    }\hfill
    \subfloat[Real-world obstacle placement]{
        \includegraphics[width=0.31\textwidth]{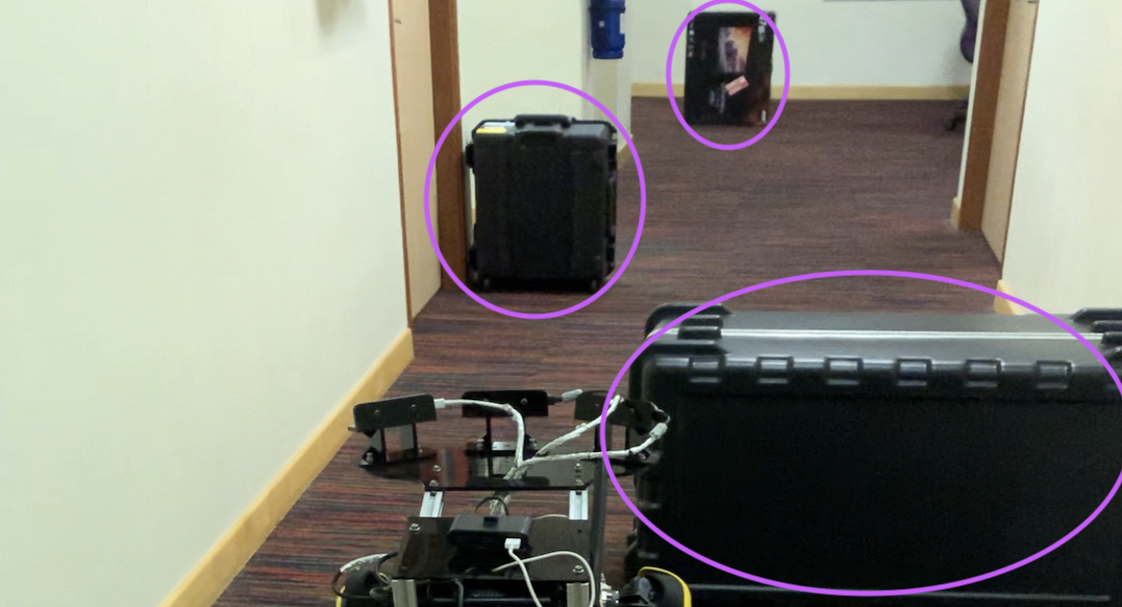}
        \label{subfig:obstacle_placement}
    }

    \caption{(\textit{a}-\textit{e}) show the modified maps used with both DEC and DWA for the navigation robustness experiments. The start and end points of the navigation task are specified by the green and red stars respectively. (\textit{f}) shows the real-world placement of obstacles that the robot has to avoid in all the navigation tasks. (\textit{a}) is the \textbf{Original} map built with Cartographer SLAM, which is both metrically accurate and captures the obstacles placed in the environment. (\textit{d}) is the same as (\textit{a}), but with the obstacles removed. (\textit{b}), (\textit{e}) are scaled versions of (\textit{d}). (\textit{b}) scales the map lengthwise from the original aspect ratio of 1.84 to 1.5, while (\textit{e}) scales it further to 1.33. Finally, (\textit{c}) is a hand-drawn sketch of the environment that only approximately captures its structure and does not contain obstacles.}
    \label{fig:slam_maps}
\end{figure*}

We answer \ref*{qn:inaccuracies} by comparing a DLM-steered \intentionnet{} system to a classical navigation system in terms of their ability to tolerate mapping and positioning inaccuracies, which serves as an indicator of their potential to \textit{scale to long-range navigation}. We do not include an LPE-based system in the tests, as it requires accurate metric information to perform well. 

In particular we test both systems' ability to tolerate 1) missing map information, and 2) metric inaccuracies in both mapping and localisation. For (1), this is achieved by manually removing obstacles from the map provided to the systems. For (2), we distort the map in various ways so as to degrade the systems' ability to position themselves with respect to the map, as well as to render their planned paths inaccurate. The tests take place in a cluttered indoor office environment. We first generate an accurate map of the test environment in the form of an occupancy grid built using Google's Cartographer SLAM~\citep{hess_cartographer}. All distorted and modified maps are manually edited variants of this initial occupancy grid.

Given an occupancy grid, both the \intentionnet{} and classical systems use Cartographer to localise on it, and ROS \texttt{move\_base} to plan paths over it. In the context of the \intentionnet{} system, this is equivalent to using a single grid-map from the local map layer to navigate. The \intentionnet{} and classical systems differ in terms of the low-level controller used, as well as in how the low-level controller is steered to follow the planned path. The \intentionnet{} system uses the DLM-steered DECISION controller and employs the intention generation module from \autoref{sec:planning_intgen} to convert the planned path to DLM intentions. On the other hand, the classical navigation system uses DWA as the low-level controller, which directly accepts the planned path as input and tries to follow it. We abbreviate the \intentionnet{} and classical navigation systems used here as DEC and DWA respectively.

As shown in \autoref{subfig:original}, the task is to navigate from the start (green star) to the end (red star) while avoiding obstacles. If the robot goes too close ($< 10$cm) to an obstacle or wall, or if the robot gets stuck in the same position for more than 5 seconds, we count the trial as a failure. We will also intervene by manually teleoperating the robot for 2-3 seconds to shift it to a different feasible position closer to the goal, then allow the robot to continue autonomously from there. We perform $N=10$ trials and present the results for these trials in \autoref{tab:nav_robustness}.


\subsubsection{Missing map information} To analyse the systems' robustness to missing map information, we edit \textit{Original} accurate map (\autoref{subfig:original}) and remove the obstacles. The modified \textit{No obstacles} map is shown in \autoref{subfig:no_obs}.

\textbf{No obstacles.} We observe that DEC maintains the same success and average intervention rate across both \textit{Original} and \textit{No obstacles} tests, without any drop in performance. This highlights that its learned obstacle avoidance is robust enough to handle unexpected obstacles not accounted for in the global plan and reactively navigate through the narrow gaps around the obstacles. DEC is more robust than DWA in avoiding collisions with missing map information, as it outperforms DWA with both a higher success rate and lower average intervention rate in the \textit{No obstacles} setting.

\subsubsection{Metric inaccuracies}
In addition to removing obstacles, we introduce metric inaccuracies into the \textit{Original} accurate occupancy grid, so as to degrade localisation performance and to render planned paths inaccurate. This is done in two ways: by scaling the occupancy grid lengthwise to change its aspect ratio, and by replacing the \textit{Original} grid with a sketched version of itself. \autoref{subfig:scaled_1.5} and \autoref{subfig:scaled_1.33} are variants of the \textit{Original} grid scaled to different extents. These tests simulate the metric inaccuracies an \intentionnet{} system might have to handle when working with floor-plans, which might capture environment structure well but might be wrongly scaled. The hand-drawn sketch of the environment (\autoref{subfig:hand_drawn}) goes beyond wrong scaling, by misrepresenting geometric details and relative distances in the environment.


\textbf{No obstacles + Scaled.} When using the inaccurate scaled maps, DEC markedly outperforms DWA. The combined errors from metric path planning and localisation due to the scaling are significant enough that DWA misses turnings and gets stuck in corners when attempting to track the commanded metric path through the test area's narrow corridors. In contrast, DEC performs much better, succeeding at all trials in the \textit{No obstacles + Scaled to 1.5 AR} tests while DWA fails to complete any. DEC does this by by capitalising on the controller's robust learned path-following abilities. The next DLM intention on the path is always issued well in advance of the next turning/junction, relying on the controller's ability to follow the corridor while avoiding obstacles until it becomes possible to make a turn. Thus, the learned controller enables robust navigation without requiring precision in triggering DLM intention changes, compensating for the metric inaccuracies in the map and from localisation and allowing DEC to complete the trials successfully. We further stress test DEC by scaling the map down even further in the \textit{No obstacles + Scaled to 1.33 AR} test in \autoref{subfig:scaled_1.33}. We omit DWA from this test since it failed to complete any trials in the easier test. At this level of scaling the combined errors from inaccurate path planning and localisation are severe (~6-8m), leading to DEC sometimes prematurely triggering the \texttt{turn-right} intention for the final turn into the corridor leading to the goal at a previous turning where a \texttt{turn-left} is needed instead, causing the robot to get stuck in a corner. Even so, DEC is still robust enough to succeed in 3 out of 5 trials for this test.

\textbf{No obstacles + Hand-drawn.} We find that DEC also outperforms DWA when using a hand-drawn sketch that has inconsistent relative distances and misrepresents geometric features in addition to removing obstacles. These contribute to inaccuracies in path planning and localisation severe enough that the robot gets stuck in 4 out of 5 trials with the DWA system. Similar to the scaled map tests, the ability of DEC to trigger DLM intention changes without needing precision in where the changes are triggered allows it to compensate for the metric inaccuracies and achieve the same high success rate and low intervention rate that it got in the \textit{Original} and \textit{No obstacles} tests.


\begin{table*}[t]
\small\sf\centering
\caption{Comparison of navigation robustness in the presence of metric inaccuracies in positioning and mapping, across DEC and DWA systems}
  \label{tab:system}
 \begin{tabular}{l c c c c c c c c}
    \toprule
    Map modifications &
    \multicolumn{2}{c}{SR/\% ($\uparrow$)} & 
    \multicolumn{2}{c}{Avg. Int. ($\downarrow$)} & 
    \multicolumn{2}{c}{Smoothness ($\downarrow$)} &
    \multicolumn{2}{c}{Avg. Ex. Time/secs ($\downarrow$)} \\
    \cmidrule(lr){2-3} \cmidrule(lr){4-5} \cmidrule(lr){6-7} \cmidrule(lr){8-9}
    & DWA & DEC & DWA & DEC & DWA & DEC & DWA & DEC \\
    \midrule
    Original & 60 & \textbf{80} & 0.6 & \textbf{0.2} & 0.042 & \textbf{0.016} & 60.9 & \textbf{36.3}\\
    No obstacles & 60 & \textbf{80} & 0.8 & \textbf{0.2} & 0.045 & \textbf{0.014} & 55.4 & \textbf{46.6} \\
    No obstacles + Scaled to 1.5 AR & 0 & \textbf{100} & 7 & \textbf{0} & \textbf{0.017} & 0.018 & 74.3 & \textbf{36.8} \\
    No obstacles + Scaled to 1.33 AR & - & \textbf{60} & - & \textbf{0.6} & -& \textbf{0.019} & - & \textbf{36.7} \\
    No obstacles + Hand-drawn & 20 & \textbf{80} & 4 & \textbf{0.2} & 0.027 & \textbf{0.018} & 78.0 & \textbf{36.0}\\ 
    
 \bottomrule
 \label{tab:nav_robustness}
\end{tabular}
\end{table*}

\section{Long-range navigation in complex environments with \inetlr{}}
\label{sec:long_range}

In this section, we deploy the proposed \inetlr{} system on realistic, long-range navigation tasks through varied and complex environments. We navigate over routes ranging from several hundred metres to a kilometre, through a mix of indoor and outdoor environments.

\subsection{Routes and experimental setup}
We select two routes through the NUS campus that 1) highlight our planner and map system's flexibility in handling a mix of indoor and outdoor environments, and 2) showcase the DECISION controller's performance across diverse areas and its ability to generalise to novel environments. \autoref{fig:long_range_routes} shows the variety of environments and terrains that the robot encounters along these routes.

Route 1 in \autoref{subfig:long_range_route1} is an approximately 1km long route through mixed indoor-outdoor environments. The route starts inside a building and ends outdoors, at a cul-de-sac on a public road. This route requires the robot to traverse between different floors of a building by taking the stairs, to transit to different buildings via connecting linkways, to walk through an outdoor park environment and along a public road. The route is specified as a series of waypoints with respect to floor-plans of the NUS buildings downloaded from the Internet. The floor-plans are minimally processed - as described in \autoref{sec:5}, we only specify connections between adjacent buildings, and roughly hand-draw outdoor paths not specified in the floor-plans. The robot plans a path over this collection of floor-plans and generates intentions from the planned path. For the entirety of this route, the robot only localises using odometry as described in \autoref{sec:5_loc}, without any form of loop closure or error correction in the robot's pose. Around 60\% of the route takes place in areas outside the controller's training dataset.

Route 2 in \autoref{subfig:long_range_route2} is a 550m long outdoor route that starts in the driveway of a building, follows the sidewalk beside the main road, passes through various junctions and turnings, goes through a carpark, and ends at the lobby of the destination building. Since the route is purely outdoors, we only use the GPS map layer of the map system. The low-level controller has not been trained on any data from the entire route, and in fact has never been trained to follow a sidewalk next to a road. Thus, this route tests the generalisation capabilities of the DECISION controller.

\inetlr{} is run on the onboard Xavier computer when executing both these routes. If the robot deviates significantly from the intended path, or if the robot's controller gets stuck at a particular spot, we intervene to correct its motion by teleoperating the robot back onto the intended path.

\subsection{Results}
\begin{figure}
    \centering
    \includegraphics[width=0.95\linewidth]{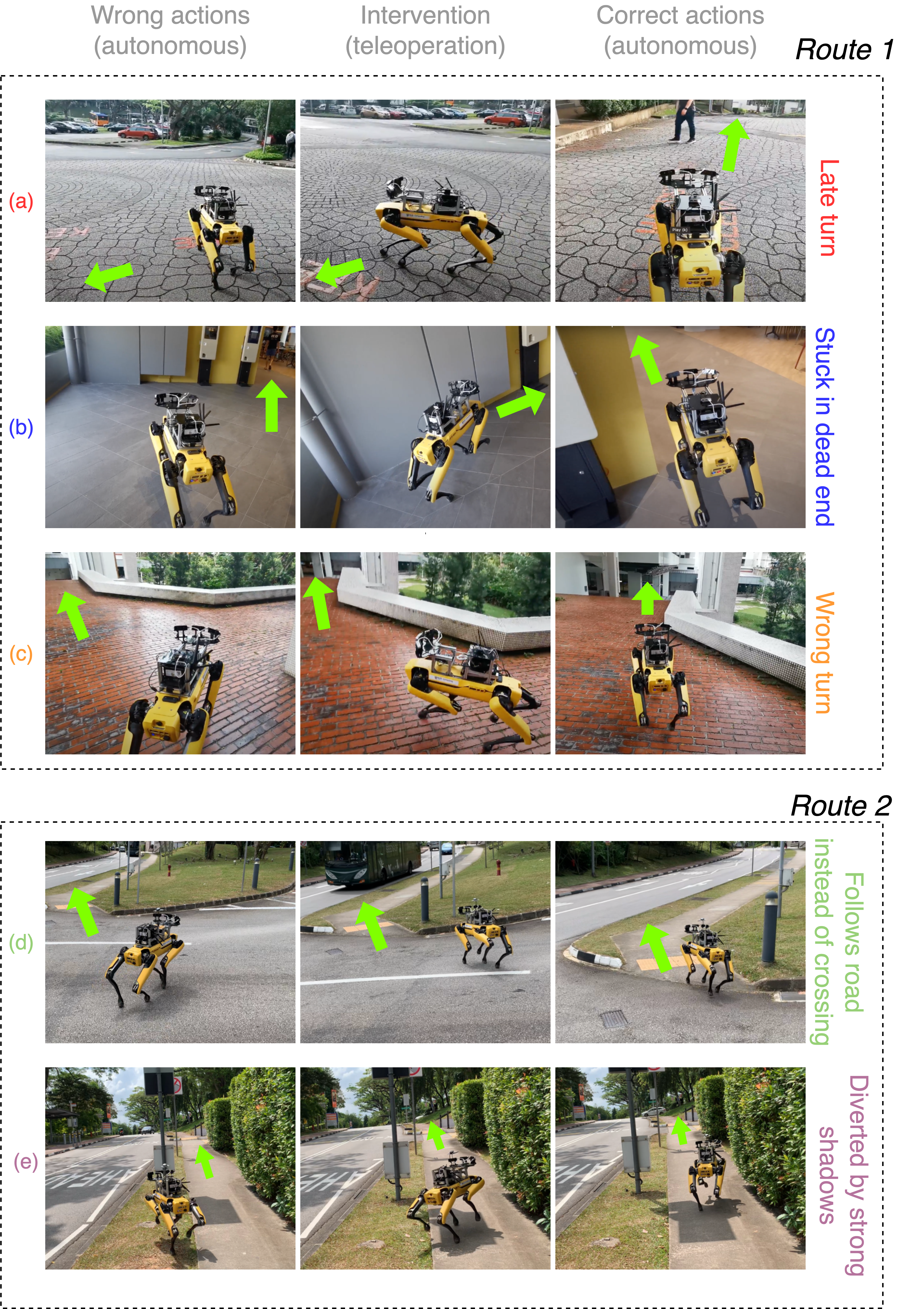}
    \caption{Each row shows a common failure mode encountered along the routes. The columns show the system's failure, the operator's intervention to recover the system, then the system resuming navigation. Across the columns, the green arrow indicates the desired direction for the robot to navigate in. \textbf{(a)} System issues a late turn near end of route due to localisation drift of $\sim$10m. We help to reorient the robot towards the side road that is its goal. \textbf{(b)} Robot misses narrow path to its right when navigating an ambiguous open space with the forward intention, getting stuck in the left corner. We orient it towards the path. \textbf{(c)} System makes a late intention change due to metric inaccuracies in the planned path, causing robot to execute its previous right intention for too long and face the wrong path. We orient the robot to face forward again. \textbf{(d)}  When crossing the road, the robot ends up following the road instead of heading to the opposite sidewalk. We orient it towards the sidewalk. \textbf{(e)} The robot is diverted onto the grass as it interprets the stark shadow from the lamp-post as the path boundary. We orient it along the path.}
    \label{fig:lr1_failure}
\end{figure}

We performed two runs of Route 1, covering the entire route each time with only 4 and 3 interventions respectively. We also performed two runs of Route 2, reaching the goal with 6 interventions on each attempt. Some of the common failure modes observed along the routes which necessitated interventions are highlighted in \autoref{fig:lr1_failure}.

We draw several conclusions from the results. Firstly, we validate that \inetlr{} is indeed effective at robustly navigating long distances with noisy metric localisation. Based on the GPS locations of the start and end points on Route 1, we estimate that the odometry drifts by around 15m in the x-y plane over the entire route. In spite of this, \inetlr{} only needed 2 interventions due to issues arising from odometry drift over the course of both the Route 1 runs. Specifically these interventions were needed because the accrued drift close to the end of the route shifted the turning points along the path significantly, leading \inetlr{} to make the turnings late. 

Secondly, we find that the DECISION controller can handle a wide range of different environments and also exhibits strong generalisation capabilities. Over the various runs of both routes 1 and 2, the robot covers in excess of 2km of novel areas unseen in its training data. \autoref{fig:long_range_routes} shows the wide range of visual appearances present across these novel environments. In spite of this, DECISION is largely able to follow paths and execute the DLM behaviours accurately in these unseen areas, with only 3-4 short interventions attributable to controller failure when navigating through these novel areas. DECISION is also able to navigate effectively through a mix of indoor and outdoor environments. 

Thirdly, we note that the DLM-based \inetlr{} system performs best in structured environments with well-defined paths and turnings that do not require highly precise manoeuvres to negotiate. We observe that \inetlr{} performs robustly when there are clear structures or features to demarcate the robot's path of travel. While DLM's coarseness means precise motions are not possible, the presence of clear structures and features can guide the controller and alleviate this issue somewhat. On the other hand, the `Stuck in dead end' example in \autoref{fig:lr1_failure} shows how the controller can fail in open areas where paths are ambiguous. While executing the \texttt{go-forward} intention here, the robot simply walks forward and misses the path on its right, getting stuck in the left hand corner of the open area. We surmise the lack of clear structure in the form of walls, textures or markings on the floor contributes to the failure. 

We also make a related observation that the DECISION controller appears to rely heavily on texture cues to discern path boundaries and follow paths, especially in areas it has no training data of. For instance, 7 out of 12 of the interventions over both runs of Route 2 occur at road crossings, where the robot attempts to follow the road instead of crossing to the opposite side. The asphalt roads usually have a distinctly different appearance and texture from the sidewalks, and the robot will usually attempt to follow along the path with the same texture during crossing, thus choosing to track the road. From these, we surmise that the controller has learnt to use texture as a strong feature in identifying paths in novel areas.

\section{Discussion}

We proposed the \intentionnet{} navigation system architecture, that combines a classical global planner with a learned low-level controller. Using classical planning algorithms are capable of compositional generalisation to distant goals, enabling us to plan large-scale paths. At the same time, we show that a low-level controller learned end-to-end is more robust than classical hand-crafted local planners, as the learning can compensate for imperfections in the system decomposition and in the models/priors used. Crucially, parameterizing the controller with a neural network grants the flexibility for goals to be specified to the controller in a wide variety of ways. We call these varied means of goal specification \textit{intentions}, and explore several different types of intentions: \textit{Local Path and Environment} (LPE) and \textit{Discretised Local Move} (DLM). Each of these intentions imbues the \intentionnet{} system with different capabilities. We note in particular that DLM-based \intentionnet{} systems can be robust to noisy metric localisation due to the coarse, discrete nature of the intention.


Building on this, we developed an instance of the \intentionnet{} system called \inetlr{}, that is designed to enable robust kilometre-scale navigation over a mix of complex indoor and outdoor environments. \inetlr{} uses DLM intentions for robustness to localisation drift over long distances, and employs the DECISION controller to robustly learn the behaviours corresponding to the DLM intentions. Our experiments show that \inetlr{} is effective at long-range navigation through diverse and challenging environments, even with only noisy and drifting odometry. 

The \inetlr{} navigation system still has several limitations. Firstly, DLM intentions are not well-suited to guiding the robot through less-structured environments like open spaces which contain no clear paths for the robot to follow. Secondly, DLM is unable to manoeuvre the robot with precision due to its coarseness. There is a large design space for intentions, and we might be able to craft better intentions to overcome these limitations, or to combine multiple different intention types within the same navigation system, using different intentions where appropriate. We leave this exploration as future work.
\bibliographystyle{plainnat}
\bibliography{references}

\clearpage

\appendix
\section{DECISION controller architecture} \label{decision:architecture}
Two technical details are important to stabilize training and to improve genearlization of our DECISION controller.

\textbf{Normalization layers.} We use a GroupNorm \citep{group_norm} layer ($G=32$) after each convolution in the cell. There are two practical reasons. Firstly, we need normalization to stabilize the gradient flow, as we observe that the gradients appear statistically unstable when they propagate through deep layers and long time horizons, which makes training much harder. Secondly, normalization reduces variance in the intermediate features. When the batch size is small, \eg{} a batch size of one at deployment time, features show more statistical variability than at training time, which destabilizes runtime performance. Among different normalizations \citep{layer_norm, group_norm, batch_norm}, we find GroupNorm \citep{group_norm} with a large $G$ the most effective.

\textbf{Dropout.} As the observations in adjacent time steps are highly correlated, we need to prevent the memory module from overfitting to the redundancy in the temporal features. We adapt 1D dropout methods for RNNs \citep{rnn_drop, recurrent_dropout, mc_dropout} to 2D features by applying it channel-wise. We find that jointly using recurrent dropout \citep{recurrent_dropout} and Monte Carlo dropout \citep{mc_dropout} yields the best real-world performance. 

\textbf{Overall structure. } Firstly, dropout is applied to the input $x_t$ and hidden state $h_{t-1}$ \citep{mc_dropout} 
\begin{align*}
    x_t = d(x_t),\:h_{t-1} = d(h_{t-1})
\end{align*}

Then, we compute the three gates $i_t$, $f_t$, and $g_t$, 
\begin{align*}
     i_t & = \sigma(GN(W_{xi} \circ x_t + W_{hi} \circ h_{t-1} + W_{ci} \circ c_{t-1} + b_i)) \\
     f_t & = \sigma(GN(W_{xf} \circ x_t + W_{hf} \circ h_{t-1} + W_{cf} \circ c_{t-1} + b_f)) \\
     g_t & = tanh(GN(W_{xc} \circ x_t + W_{hc} \circ h_{t-1} + b_c))
\end{align*}

In addition, dropout is applied to the cell update $g_t$ to alleviate overfitting to redundant temporal cues \citep{recurrent_dropout}
$$
    g_t = d(g_t)
$$
$$
    c_t = f_t * c_{t-1} + i_t * g_t  
$$
Finally, we update and output the hidden state  
$$
    o_t = \sigma(GN(W_{xo} \circ x_t + W_{ho} \circ h_{t-1} + W_{co} \circ c_t + b_o)) 
$$
$$
    h_t = o_t * tanh(c_t) 
$$
where $d$ is the dropout operation, $W$ represents model weights, $\circ$ denotes convolution, $*$ is the Hadamard product, and $GN$ denotes GroupNorm. In this way, history is encoded recursively into $c_t$ across time steps, which is used to generate an enriched representation $h_t$ from the input $x_t$.

\section{DECISION controller training} \label{decision:training}

Here we provide more details on dataset post-processing and model training. 

\textbf{Dataset balancing. } The frequency of occurrences of different intentions is imbalanced in our dataset. We find that it is crucial to learn from a balanced mix of different intentions to prevent the controller from learning only the dominant behaviour - this tends to be the \intentforward{} behaviour, and such a controller may never make turns or recover from deviations. In practice, we subsample the dataset to balance the intention occurrence frequencies, and resample the data at each epoch to increase diversity of samples observed during training.


\textbf{Training. } We train \proposed{} with Truncated Backpropagation Through Time (TBPTT) \citep{tbptt} with AdamW optimizer \citep{adamw}. To reduce temporal redundancy, we take every third frame in the dataset to construct model input sequences of length $L=35$. The data is then normalized with ImageNet \citep{imagenet} statistics and augmented with strong color jittering. In each training iteration, the model predicts controls for $k_1=5$ observations, and backpropagation is done every $k_2=10$ predictions.

Drawing from the linear scaling rule \citep{imagenet_1hr, generalization_sgd}, we define our learning rate $LR$ as linear in batch size $BS$ and $k_2$, \ie, 
$$LR = BaseLR * BS * k_2$$ 
where $BaseLR$ and $BS$ are set to 1e-7 and 36 respectively. We use a weight decay rate of 5e-4 and a dropout rate of 0.3. Input images are resized to 112$\times$112 at both training and deployment time for fast computation. We train the model for 200 epochs and decay the learning rate at the 70th and 140th epochs. Training takes $\sim 71$ hours with $4\times$ RTX2080Ti. We reserve 10\% of the data for training-time validation, though we found out that offline loss values only very weakly correlated with online performance, similar to the finding by \cite{offline_eval_il}. Thus, we rely on real-world testing for hyperparameter tuning.

\section{Map construction} \label{appendix:mapbuilding}

The map construction proceeds in several steps: first we construct the GPS map layer, then we identify areas not covered in the GPS map layer like indoor areas and construct the local map layer such that it covers these areas, then finally we define the connectivity graph over the layers. While there are well-established methods for constructing the GPS and local map layers from \textit{direct experience} of the environment, doing this can be effortful and time-consuming. We highlight that with the right design choices, the \intentionnet{} architecture can allow us to sidestep this problem and construct the map system \textit{from commonly available resources}. 

In particular, the \inetlr{} system can make use of commonly available representations in \textit{both its GPS and local map layers}. GPS and outdoor traversability data is widely available and easy to access, allowing easy construction of the GPS map layer. In addition, the local map layer can make use of commonly available representations like visitor floor plans. Even though such representations may contain incomplete environment information or metric inaccuracies, the controller's robust local obstacle avoidance capabilities as well as the use of DLM intention provides robustness against these issues.

We discuss map construction for the \inetlr{} system specifically. For a given area, we can construct the GPS map layer by downloading geo-referenced road network data from OpenStreetMap~\citep{osm}. The data from OpenStreetMap specifies the contours of roads and walking routes, and also identifies the locations of key buildings and structures in the environment.

Next, we identify areas of interest that we want the robot to be capable of navigating in, but which are not represented in the GPS map layer. In the urban areas we test in, these are usually indoor areas inside buildings. To obtain the occupancy grids for these indoor areas, we download their floor plans from the university's website and simply binarize them to convert them into an occupancy grid format.

Finally, we define the connectivity graph over the map layers by manually annotating \textit{Exit} nodes and connecting them up. On each map in the local map layer, we add \textit{Exit} nodes at all stairs, linkways and entrances of the building. We then add the appropriate inter-layer and inter-grid edges between pairs of mutually reachable \textit{Exit} nodes, and then add intra-grid edges between all pairs of \textit{Exit} nodes on the same map.

\end{document}